\documentclass[runningheads]{llncs}

\usepackage{eccv}

\usepackage{eccvabbrv}

\usepackage{graphicx}
\usepackage{booktabs}

\usepackage[accsupp]{axessibility}  % Improves PDF readability for those with disabilities.

\usepackage{hyperref}

\usepackage{orcidlink}
\usepackage{algorithm}
\usepackage{algorithmic}
\usepackage{pifont}  

\newcommand{\cmarkred}{\textcolor{red}{\ding{51}}}
\usepackage{caption}
\usepackage{amsmath} 
\usepackage{amsmath, amssymb}
\usepackage{booktabs}
\usepackage[table,xcdraw]{xcolor}
\definecolor{lightgray}{gray}{0.9}
\usepackage{multirow} 
\usepackage{booktabs}
\usepackage{wrapfig}
\usepackage{tikz}
\usepackage{graphicx}
\usepackage{subcaption}
\usepackage{listings}

\usepackage{svg}

\definecolor{codebg}{HTML}{F7F7F7}
\definecolor{codestring}{RGB}{196,26,22}
\definecolor{codecomment}{RGB}{0,116,0}
\definecolor{codekeyword}{RGB}{170,13,145}
\definecolor{codeclass}{RGB}{38,38,255}

\lstdefinestyle{pythonstyle}{
  backgroundcolor=\color{codebg},
  language=Python,
  basicstyle=\ttfamily\small,
  keywordstyle=\color{codekeyword}\bfseries,
  stringstyle=\color{codestring},
  commentstyle=\color{codecomment},
  classoffset=1,
  morekeywords={nn, Module},
  keywordstyle=\color{codeclass},
  classoffset=0,
  frame=single,
  rulecolor=\color{gray},
  showstringspaces=false,
  tabsize=2,
  breaklines=true
}
\usepackage{caption}
\usepackage{xr}
\usepackage{hyperref}

\begin{document}

\title{From Blur to Detail: Frequency-Decoupled Latent Optimization
for Realistic Image Generation} 
\title{Decoupling High and Low Frequencies for Faithful Image Generation with Fine Details}

\titlerunning{DeBaT}

\author{Tejaswini Medi\inst{1}\orcidlink{0009-0004-6345-2693
}\and
Hsien-Yi Wang\inst{1} \and
Arianna Rampini\inst{2} \and
Margret Keuper\inst{1,3}\orcidlink{0000-0002-8437-7993}}

\authorrunning{T.~Medi et al.}

\institute{University of Mannheim, Germany \and
Autodesk AI Lab \and
MPI for Informatics, Saarland Informatics Campus, Saarbrücken,  Germany \\
\email{\{tejaswini.medi,keuper\}@uni-mannheim.de}}

\maketitle

\begin{abstract}
Latent generative models compress images into learned embeddings prior to synthesis, and the generation quality critically depends on how faithfully these embeddings preserve visual detail. We observe that while such embeddings are effective at reconstructing low frequency structure, they struggle to recover sharp high frequency details that are essential for perceptual realism. Conventional reconstruction objectives implicitly prioritize coarse structural information over high frequency content, which can lead to overly smoothed outputs and degraded visual quality in textured regions. 
Motivated by this observation, we propose DeBaT, a \underline{De}coupled frequency \underline{Ba}nd \underline{T}okenizer that explicitly separates the learning of low and high frequency band embeddings. This decoupling enables accurate reconstruction of fine details while preserving global coherence. Integrated into a latent diffusion based generative model, DeBaT allows for sharper and more realistic samples than previous latent tokenizers, confirming that the explicit decoupling of high and low frequency bands eases the preservation of visual details in learned embedding spaces.
%We integrate our tokenizer into a latent diffusion based generative model to produce sharper and more realistic samples. Experimental results show that frequency aware optimization improves both reconstruction fidelity and generative quality.

  
\end{abstract}

\section{Introduction}

\begin{figure}[t]
\vspace{-8pt}
\centering
\setlength{\tabcolsep}{1pt}
\renewcommand{\arraystretch}{0.9}
\resizebox{0.8\textwidth}{!}{
\begin{tabular}{@{}c@{\hspace{5mm}}ccc@{\hspace{5mm}}ccc@{}}

% ---------------- ORIGINALS ----------------
&\small{\textbf{original}}&\small{\textbf{VA-VAE}} &
\small{\textbf{DeBaT (ours)}} 
&\small{\textbf{original}}&\small{\textbf{VA-VAE}} &
\small{\textbf{DeBaT(ours)}} \\
\begin{tikzpicture}
\node[anchor=south west,inner sep=0] at (0,0)
{\includegraphics[width=0.25\columnwidth]{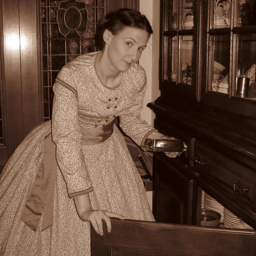}};
\draw[red, ultra thick](1.1,2.1) rectangle (1.9,3.0);
\draw[red, ultra thick](1.6,1.03) rectangle (2.35,1.73);
\end{tikzpicture}
&
\includegraphics[width=0.25\columnwidth,trim={3.29cm 6.4cm 3.4cm 0.29cm},clip]{teaser/I1orig.png}
&
\begin{tikzpicture}
\node[anchor=south west,inner sep=0] at (0,0)
{\includegraphics[width=0.25\columnwidth,trim={3.29cm 6.4cm 3.4cm 0.29cm},clip]{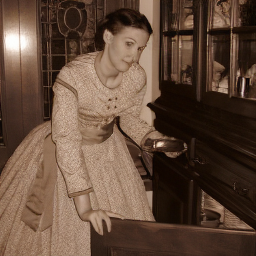}};
\draw[-to,shorten >=-1pt,red,ultra thick] (0.9,2.1) -- (1.3,1.7);
\end{tikzpicture}
%&\includegraphics[width=0.25\columnwidth,trim={3.29cm 6.4cm 3.4cm 0.29cm},clip]{teaser/30071_recon_litevae.png}
&
\includegraphics[width=0.25\columnwidth,trim={3.29cm 6.4cm 3.4cm 0.29cm},clip]{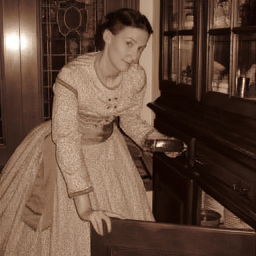}
&
\includegraphics[width=0.25\columnwidth,trim={4.79cm 3.4cm 2.4cm 3.79cm},clip]{teaser/I1orig.png}
&
\begin{tikzpicture}
\node[anchor=south west,inner sep=0] at (0,0)
{\includegraphics[width=0.25\columnwidth,trim={4.79cm 3.4cm 2.4cm 3.79cm},clip]{teaser/I1vavae.png}};
\draw[-to,shorten >=-1pt,red,ultra thick] (1.4,1.8) -- (1.1,1.3);
\end{tikzpicture}
%&\includegraphics[width=0.25\columnwidth,trim={3.29cm 6.4cm 3.4cm 0.29cm},clip]{teaser/30071_recon_litevae.png}
&
\includegraphics[width=0.25\columnwidth,trim={4.79cm 3.4cm 2.4cm 3.79cm},clip]{teaser/recon_30071_ours.png}%teaser/I1wavae.png}
\\

\begin{tikzpicture}
\node[anchor=south west,inner sep=0] at (0,0)
{\includegraphics[width=0.25\columnwidth]{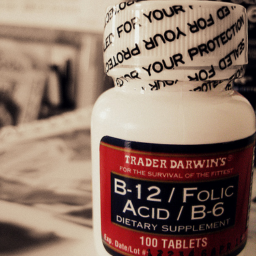}};
\draw[red, ultra thick](1.1,2.85) rectangle (1.91,2.065);
\draw[red, ultra thick](1.6,0.01) rectangle (2.4,0.9);
\end{tikzpicture}
&
\includegraphics[width=0.25\columnwidth,trim={3.29cm 6.4cm 3.4cm 0.29cm},clip]{teaser/I2orig.png}
&
% ---------------- TOP ZOOM ----------------
\begin{tikzpicture}
\node[anchor=south west,inner sep=0] at (0,0)
{\includegraphics[width=0.25\columnwidth,trim={3.29cm 6.4cm 3.4cm 0.29cm},clip]{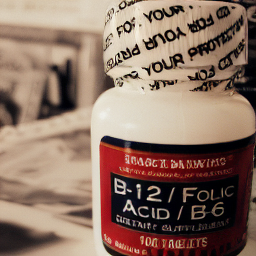}};
\draw[-to,shorten >=-1pt,red,ultra thick] (1.0,1.4) -- (0.99,0.9);
\end{tikzpicture}
%&\includegraphics[width=0.25\columnwidth,trim={3.29cm 6.4cm 3.4cm 0.29cm},clip]{teaser/36028_recon_litevae.png}
&
\includegraphics[width=0.25\columnwidth,trim={3.29cm 6.4cm 3.4cm 0.29cm},clip]{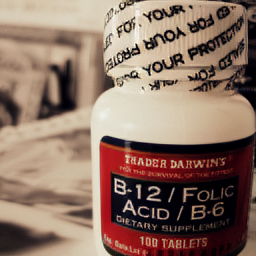}
&
\includegraphics[width=0.25\columnwidth,trim={4.79cm 0.0cm 1.9cm 6.69cm},clip]{teaser/I2orig.png}
&
\begin{tikzpicture}
\node[anchor=south west,inner sep=0] at (0,0)
{\includegraphics[width=0.25\columnwidth,trim={4.79cm 0.0cm 1.9cm 6.69cm},clip]{teaser/I2vavae.png}};
\draw[-to,shorten >=-1pt,red,ultra thick] (1.3,2.1) -- (1.3,1.5);
\end{tikzpicture}
%&
%\includegraphics[width=0.25\columnwidth,trim={4.79cm 0.0cm 1.9cm 6.69cm},clip]{teaser/36028_recon_litevae.png}
&\includegraphics[width=0.25\columnwidth,trim={4.79cm 0.0cm 1.9cm 6.69cm},clip]{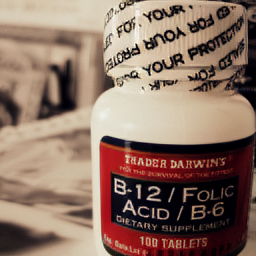}%teaser/I2debat.png}

\end{tabular}
}

\vspace{-6pt}
\caption{ Reconstruction quality for two examples. We compare the reconstruction of VA-VAE to our approach DeBaT. Highlighted regions emphasize textures and edges. DeBaT better preserves high-frequency details and sharp structures, producing reconstructions visually closer to the input. Arrows indicate reconstruction errors.}
\label{fig:teaser_main}
\vspace{-12pt}
\end{figure}

Latent generative modeling~\cite{Blattmann2023_VideoLDM, Dong2024_GPLD3D, Luo2024_LCM, Zhou2024_LGD} has become a central paradigm in modern content generation, enabling the synthesis of high fidelity visual data. These models typically operate in compressed latent spaces learned through variational autoencoders such as VAEs~\cite{kingma2013auto, li2024autoregressive}, where generation quality depends critically on the expressiveness and reconstruction fidelity of the latent representations. Increasing latent dimensionality can enhance representational capacity, but often yields diminishing visual generation improvements together with significantly higher computational cost. VAVAE~\cite{yao2025vavae} highlights this as an optimization dilemma inherent in latent modeling: pixel level reconstruction objectives favor faithful reconstruction as capacity increases, whereas generative objectives require semantically structured latent spaces. By aligning latents with pretrained vision foundation models, VAVAE improves convergence and semantic realism. However, despite these advances, VAVAE often lacks to faithfully reconstruct %reconstructed outputs frequently lack 
sharp textures and fine details, particularly in regions dominated by high frequency content such as text, as shown in ~\autoref{fig:teaser_main}. %This limitation restricts perceptual realism. 
In this work, we propose a simple yet effective approach to learn detail preserving latent embeddings, Decoupled frequency Band Tokenizer (DeBaT). %While prior works

%%%%%%%%%%%%%%%%%%%%%%%%%%%%%%%%%%%%%%%%%%%%%%%%%%%%%%%%%%%%%%%%%%%%%%%%%%%%%%%%%%%%%%%% Fig 1 %%%%%%%%%%%%%%%%%%%%%%%%%%%%%%%%

Our proposed approach is motivated by a comprehensive frequency based analysis of the reconstruction behavior of state-of-the-art latent tokenizers, which serve as the foundation of modern latent generative pipelines~\cite{yao2025vavae, li2024autoregressive, tian2024visual, xie2024sana, xie2025sana}. For this analysis, we focus on latent diffusion and autoregressive models, which represent the current frontier of high quality visual generation. These methods typically follow a two stage design: a latent tokenizer, often implemented as a VAE variant, learns quantized or continuous embeddings, and a probabilistic generative model is subsequently trained within this latent space.

Our spectral analysis uncovers a consistent limitation. While different tokenizers vary in their ability to reconstruct low frequency structure, their high frequency reconstruction error remains strikingly similar across architectures, \ie they become ever better at encoding semantic properties and global coherence. At the same time, high frequency components, which encode perceptually critical information such as textures and fine structural details and are therefore equally important for perceptual realism, are systematically under-reconstructed. % compared to dominant low frequency information. %As a result, improvements in low frequency fidelity primarily enhance global coherence and quantitative metrics, whereas accurate high frequency reconstruction is essential for perceptual realism. 
This reveals a fundamental imbalance in current latent tokenizers: they are optimized for structural faithfulness but not for fine detail preservation. %An ideal latent representation must simultaneously maintain global structure and preserve fine details.

%%%%%%%%%%%%%%%%%%%%%%%%%%%%%%%%%%%%%%%%%%%%%%%%%%%%%%%%%%%%%%%%%%%%%%%%%%%%%%%%%%%%%%%%%%%%%%%%%%%%%%%%%%%%%%%%%%%%%%%%%%%%%%%%
To faithfully encode fine details, \ie high-frequency information, our proposed DeBaT (Decoupled frequency Band Tokenizer)
explicitly separates and independently optimizes low and high frequency components via wavelet decomposition. Unlike conventional tokenizers that operate purely in the spatial domain, DeBaT performs an explicit frequency band decomposition and learns distinct latent embeddings for low and high frequency components. 
This enables faithful reconstruction of fine textures while preserving global structural coherence. ~\autoref{fig:teaser_main} illustrates a visual comparison with VA-VAE, demonstrating improved preservation of sharp structures and high frequency details. By integrating DeBaT into a lightweight latent diffusion model, LightningDiT, we observe that improved spectral reconstruction directly translates into high quality realistic generative samples.

\vspace{0.5cm}

\noindent\textbf{Our main contributions are summarized as follows:}
\begin{itemize}
    \item[\cmarkred] We present a comprehensive frequency-based reconstruction evaluation of latent tokenizers used in modern generative pipelines. Our analysis reveals that improvements in overall reconstruction performance are primarily driven by enhanced low-frequency reconstruction, while high-frequency details remain under-preserved across both spatial and spectral latent tokenizers.

    \item[\cmarkred] We propose DeBaT, a Decoupled frequency Band Tokenizer that explicitly separates and independently optimizes low- and high-frequency components via wavelet decomposition. By mitigating spectral optimization coupling, DeBaT learns frequency-aware and more expressive latent representations that preserve both global structure and fine details.

    \item[\cmarkred] We integrate DeBaT into LightningDiT and demonstrate improved generative quality on ImageNet. Our results highlight the importance of frequency-aware latent modeling for achieving sharper and more perceptually realistic image reconstruction and synthesis.
\end{itemize}

\section{Related Work}

\subsection{Latent Embeddings for Image Generation}

Latent representation learning models, or visual tokenizers, are widely used to learn compressed latent representations that capture high-dimensional visual data and enable efficient generation. Encoder-Decoder frameworks such as VAEs~\cite{kingma2013auto} and their variants form the foundation for learning these latent representations, which are vital for modeling latent generative models. Notably, Latent Diffusion Models (LDMs)~\cite{rombach2022high} operate over learned latent embeddings to facilitate high-quality synthesis, highlighting the importance of effective latent space modeling. Discrete tokenization techniques, such as VQ-VAE~\cite{van2017neural,10.1007/978-3-031-54605-1_10}, convert images into latent code sequences that can be effectively modeled by transformers. VQGAN~\cite{esser2021taming} further improves this framework by incorporating adversarial losses to learn perceptually meaningful tokens, while RQ-VAE~\cite{lee2022autoregressive} enhances the expressiveness of embeddings through stacked residual quantizers. Recent methods such as TiTok~\cite{yu2024image}, FlexTok~\cite{bachmann2025flextok}, and OneDPiece~\cite{miwa2025one} explore compact and adaptive tokenization strategies, aiming for better compression and controllable generation quality.

To enhance reconstruction quality of visual tokenizers, hierarchical and regularized latent representation learning models have been proposed. NVAE~\cite{vahdat2020nvae} introduces spectral regularization within residual hierarchies for stable training, while HQ-VAE~\cite{takida2023hq} and VAR~\cite{tian2024visual} use multi-scale vector quantization to better model local and global structure in latent spaces. EQ-VAE~\cite{kouzelis2025eq} imposes equivariance constraints for robustness to geometric changes. Additionally, frequency-aware losses such as Fourier-based regularization~\cite{bjork2022simpler} and scale-equivariant decoding~\cite{skorokhodov2025improving} have been used to control the spectral properties of the latent reconstructions. A parallel direction focuses on aligning latent spaces with external vision encoders. For example, REPA~\cite{repa} introduces denoising-time regularization based on foundation model features to accelerate convergence. VA-VAE~\cite{yao2025vavae} builds on this by jointly aligning tokenizer latents with pretrained features and optimizing for downstream generation quality. ReaLS~\cite{xu2025exploring} introduces semantic priors directly into the latent space to improve representation capacity. Our work builds on VA-VAE but identifies that such alignment alone does not sufficiently preserve high frequency fine details. We propose an explicit decoupling and optimization of low- and high-frequencies during tokenization, leading to improved fidelity in both latent space representation, and by consequence in reconstruction and generation quality.

\subsection{High-Frequency Content in Image Reconstruction \& Synthesis}
High-frequency content is essential for perceptual realism but challenging to reconstruct, particularly for diffusion-based models. Prior analyses show that frequency components are not reconstructed uniformly across the denoising trajectory~\cite{yang2023diffusion}, resulting in a loss of textures and sharp structures. To address this, various frequency-aware approaches have been developed across generative paradigms. In GANs, SWAGAN~\cite{gal2021swagan} and WaveFill~\cite{yu2021wavefill} integrate wavelet transforms into their architectures to improve generation fidelity. Focal frequency loss~\cite{jiang2021focal} and WINE~\cite{kim2025wine} address frequency imbalance by reweighting reconstruction errors and correcting low-frequency bias in inversion, respectively. Diffusion models have incorporated frequency priors at both training and inference time. WaveDiff~\cite{phung2023wavediff}, Spatial-Frequency UNet~\cite{yuan2023spatial}, and spectral diffusion approaches~\cite{yang2023diffusion,guth2022wsgm} integrate wavelet-domain frequency components or gating mechanisms into the network to better optimize the spectral content. Training-free techniques like DiffuseHigh~\cite{kim2025diffusehigh} and frequency aggregation~\cite{qian2024boosting} inject wavelet-based frequency priors during inference for improved detail preservation. Additional strategies include replacing convolutions with filtered variants in FouriScale~\cite{huang2024fouriscale}, and autoregressive frequency modeling in NFIG~\cite{huang2025nfig}. Adaptations also exist for domain-specific tasks such as high-resolution~\cite{zhang2025diffusion}, underwater~\cite{zhao2024wavelet}, or low-light image generation~\cite{jiang2023low}. Beyond images, wavelet representations have further proven useful in 3D generation with diffusion and autoregressive models~\cite{hui2024make, medi20243d}, suggesting the utility of wavelet decompositions as a general tool for capturing fine-scale structure across visual modalities. While effective, these approaches typically introduce additional architectural components or post-processing steps to compensate for frequency degradation. 

\begin{figure}
    \centering
    \begin{tabular}{@{}c@{\hspace{0.5cm}}c@{\hspace{0.5cm}}c@{}}
        \scriptsize LiteVAE& \scriptsize WF-VAE& \scriptsize DeBaT (ours)\\\includegraphics[height=4cm]{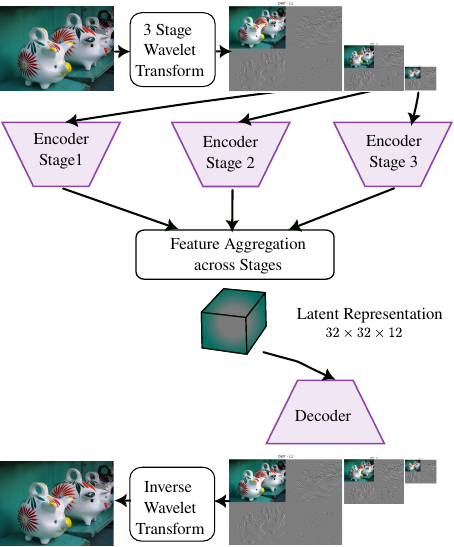}&\includegraphics[height=4cm]{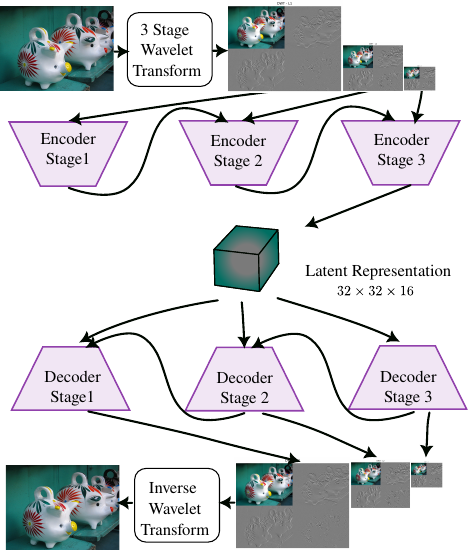}&
        \includegraphics[height=4cm]{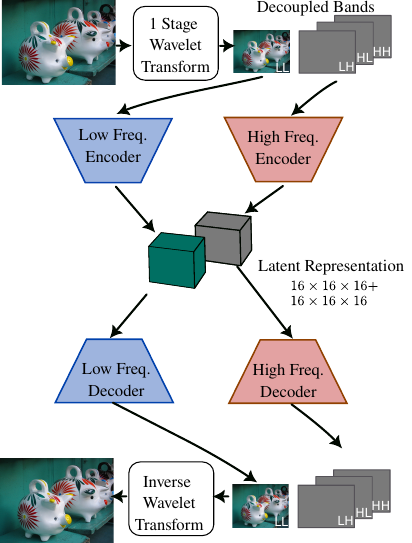}\\
        \scriptsize multiple stages, coupled bands&\scriptsize multiple stages, coupled bands&\scriptsize  single stage, decoupled bands
    \end{tabular}
    \caption{Comparison of wavelet based latent VAE tokenizer designs. Previous works for image (VA-VAE) and video (WF-VAE) generation leverage spatial compression properties of multiple wavelet stages, while low and high frequency bands are coupled in a joint latent code. We propose DeBaT, which preserves fine details by learning explicit latent codes for low and for high frequency bands through decoupled VAEs. %  Decoupled Band Tokenizer, where the low and high frequencies are decoupled.
    }
    \label{fig:method_comparison}
\end{figure}
Prior wavelet based decomposition approaches such as WF-VAE~\cite{li2025wfvae} (for video generation) and LiteVAE~\cite{sadat2024litevae} incorporate wavelet transforms into the encoding and decoding process to enhance multi scale feature extraction and improve reconstruction efficiency. Leveraging the compression properties of wavelets, these methods optimize frequency components jointly within a shared latent space and under a unified objective. As a result, they inherit the same frequency optimization coupling observed in spatial domain tokenizers, where low frequency components dominate over the learning of fine detail. In contrast, our approach DeBaT explicitly decouples the learning dynamics of low and high frequency bands by allocating separate latent embeddings and optimizing them independently prior to fusion.  ~\autoref{fig:method_comparison} compares DeBaT to different frequency based latent tokenizer models that use wavelet decomposition.
By modeling and optimizing frequency subbands (low and high) separately during tokenization, we improve the frequency aware expressivity in the latent embeddings. These latent embeddings improve both reconstruction fidelity and downstream synthesis quality when integrated with the latent generative pipeline without requiring any architectural changes at generation time.

\section{Method}

\begin{figure}[t]
    \centering
    \includegraphics[width=0.99\linewidth]{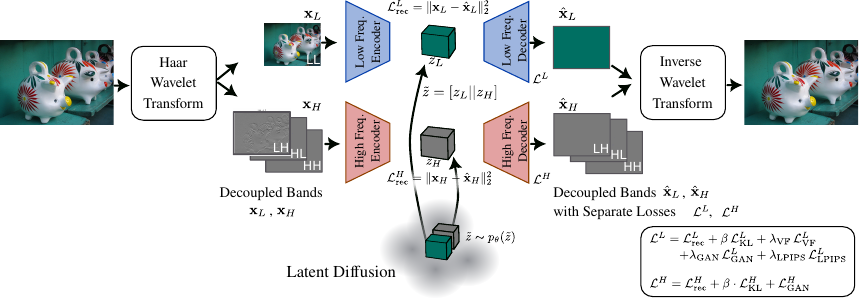}\vspace{-0.2cm}
    \caption{Overview of the proposed \textbf{DeBaT} framework. An input image $x$ is first decomposed using a discrete wavelet transform into low- and high-frequency components. Each frequency band is processed independently using dedicated encoder–decoder pairs $(E_L,D_L)$ and $(E_H,D_H)$ to learn specialized latent embeddings that capture global structure and fine details, respectively. The reconstructed frequency components are then fused through an inverse wavelet transform to obtain the final image $\hat{x}$.}
    \label{fig:debat}
\end{figure}

In this section, we present the key components of our approach. 
First, we conduct a frequency-based reconstruction evaluation of latent embeddings used in modern latent image generative pipelines. Our analysis shows that improvements in reconstruction performance primarily arise from better modeling of low-frequency structures, while high-frequency components remain comparatively underrepresented. This imbalance indicates that latent tokenizers jointly optimized over both low- and high-frequency information tend to bias the optimization toward low-frequency content, often resulting in the loss of fine-grained image details. Second, we introduce DeBaT, a novel framework that leverages wavelet decomposition to decouple latent representations into low- and high-frequency components, enabling them to be optimized independently. 
Third, we propose a simple yet effective fusion strategy that combines these frequency-aware representations into a unified latent embedding, which is subsequently integrated into the lightweight LightningDiT pipeline~\cite{yao2025vavae}. 
This design enables improved image generation by explicitly modeling frequencies in the latent space. 
An overview of the proposed DeBaT framework is shown in Figure~\ref{fig:debat}.

\subsection{Frequency Reconstruction Evaluation of Latent Tokenizers} \label{table1_metrics}

In most latent generative models, the first stage involves learning compact latent representations using visual tokenizers such as VAEs or VQ-VAEs and their architectural variants. These tokenizers learn latent embeddings \( \mathbf{z} \in \mathcal{Z} \) that encode the underlying structure of the input data \( \mathbf{x} \in \mathcal{X} \). A common measure of the representational quality of these latent embeddings is their reconstruction performance, typically evaluated using mean squared error (MSE) or the \( \ell_2 \)-reconstruction loss. The reconstruction objective is defined as:
\[
\mathcal{L}_{\text{rec}} = \|\mathbf{x} - \hat{\mathbf{x}}\|_2^2,
\]
where \( \hat{\mathbf{x}} = D(E(\mathbf{x})) \) denotes the reconstructed image obtained by encoding the input \( \mathbf{x} \) into a latent embedding \( \mathbf{z} = E(\mathbf{x}) \) with encoder \( E \) and subsequently decoding it using the decoder \( D \). While this objective provides a measure of overall reconstruction quality, it does not explicitly evaluate how well different frequency components of the image are reconstructed. In particular, the reconstruction error does not distinguish between low-frequency structural information and high-frequency details such as edges and textures. As a result, improvements in reconstruction performance may primarily reflect better recovery of dominant low-frequency content, while the fidelity of high-frequency components remains underexplored. To assess the fidelity of reconstructions across frequency bands, we perform a frequency-aware reconstruction analysis using the discrete wavelet transform (DWT) with a Haar wavelet filter. Specifically, we decompose both the input image and its reconstruction into low- and high-frequency representations. Let \( \mathcal{W}(\cdot) \) denote the wavelet decomposition operator:
\[
\mathcal{W}(\mathbf{x}) = (\mathbf{x}_L, \mathbf{x}_H), \quad 
\mathcal{W}(\hat{\mathbf{x}}) = (\hat{\mathbf{x}}_L, \hat{\mathbf{x}}_H),
\]
where \( \mathbf{x}_L \) and \( \hat{\mathbf{x}}_L \) correspond to the low-frequency (approximation) components, while \( \mathbf{x}_H \) and \( \hat{\mathbf{x}}_H \) correspond to the high-frequency (detail) components of the input and reconstructed images, respectively. We then compute frequency-specific reconstruction losses as follows:

\[
\mathcal{L}_{L} = \|\mathbf{x}_L - \hat{\mathbf{x}}_L\|_2^2, 
\qquad 
\mathcal{L}_{H} = \|\mathbf{x}_H - \hat{\mathbf{x}}_H\|_2^2.
\]

These metrics provide a more fine-grained evaluation of latent representations by measuring their ability to reconstruct structural information and fine details independently. Unlike conventional reconstruction objectives that aggregate errors across all spatial frequencies into a single pixel-space metric, our analysis explicitly separates the contribution of low- and high-frequency components. This allows us to reveal frequency-specific reconstruction in latent tokenizers that remain hidden under standard pixel based reconstruction losses. In particular, it enables a more principled understanding of whether improvements in reconstruction quality arise from better modeling of global structures or from enhanced preservation of high-frequency details.

Beyond pixel-wise reconstruction loss, we also evaluate perceptual quality using the Learned Perceptual Image Patch Similarity (LPIPS) metric~\cite{zhang2018unreasonable}. LPIPS measures perceptual similarity between input and reconstructed images using deep feature representations and better reflects human visual perception than pixel-space metrics. To further quantify the overall quality of latent reconstructions, we compute the reconstruction Fréchet Inception Distance (rFID)~\cite{heusel2017gans}. This metric compares the distributions of original and reconstructed images in the feature space of a pretrained Inception network~\cite{szegedy2016rethinking}.

\[
\text{rFID} = \|\mu_{\mathbf{x}} - \mu_{\hat{\mathbf{x}}}\|_2^2 + 
\text{Tr}\!\left( \Sigma_{\mathbf{x}} + \Sigma_{\hat{\mathbf{x}}} - 
2\left( \Sigma_{\mathbf{x}} \Sigma_{\hat{\mathbf{x}}} \right)^{1/2} \right),
\]

where \( (\mu_{\mathbf{x}}, \Sigma_{\mathbf{x}}) \) and \( (\mu_{\hat{\mathbf{x}}}, \Sigma_{\hat{\mathbf{x}}}) \) denote the mean and covariance of Inception features extracted from the original and reconstructed image distributions, respectively. Together, the frequency-aware reconstruction losses \( (\mathcal{L}_L, \mathcal{L}_H) \), perceptual similarity \( \mathcal{L}_{\text{LPIPS}} \), and distributional alignment measured by rFID provide a comprehensive evaluation of reconstruction quality. In particular, this combination enables us to analyze both the perceptual fidelity and the frequency-specific reconstruction behavior of latent embeddings.

\subsection{Decoupled Band Tokenizer (DeBaT)}
To enable frequency-aware optimization in latent tokenizers, we extend the standard VAE-based tokenizer by operating directly on frequency-decomposed representations instead of pixel-space inputs. Specifically, we introduce \textbf{DeBaT}, a Decoupled Band Tokenizer that explicitly separates low- and high-frequency components of the input image and learns dedicated latent embeddings for each band. Given an input image \( \mathbf{x} \in \mathcal{X} \), we apply a discrete wavelet transform \( \mathcal{W}(\cdot) \) using the Haar filter to obtain its low and high-frequency representations. $\mathcal{W}(\mathbf{x}) = \bigl( \mathbf{x}_L, \mathbf{x}_H \bigr),$ where \( \mathbf{x}_L \) and \( \mathbf{x}_H \) denote the low- and high-frequency components, respectively. We adopt the normalization strategy proposed in~\cite{mulcahy1997image} to normalize both components prior to encoding.

To model low and high frequency specific information, we employ two independent encoder-decoder pairs, \( (E_L, D_L) \) and \( (E_H, D_H) \), each responsible for learning latent embeddings $(\mathbf{z}_L, \mathbf{z}_H)$ corresponding to the low- and high-frequency inputs. This allows to capture the full frequency spectrum more effectively than standard VAEs, where low and high spatial frequencies are optimized jointly. To ensure a fair comparison with baseline VAEs, we allocate the latent dimensions of the low and high frequencies such that the concatenated latent vector has the same dimensionality as that of the standard baseline VAE, i.e. $16\times 16\times 32$ (see \autoref{fig:method_comparison}),
\[
\mathbf{z}_L = E_L(\mathbf{x}_L), \quad \mathbf{z}_H = E_H(\mathbf{x}_H),
\]
where $\mathbf{z}_L$ and $\mathbf{z}_H$ represents the low and high frequency latent embeddings.

\vspace{0.2cm}

\noindent\textbf{Low Frequency Objective.} 
To learn latent embeddings that capture the global structural information of images, we optimize the low-frequency component using the VAVAE training objective~\cite{yao2025vavae}. VAVAE has demonstrated strong performance among latent tokenizers; it improves reconstruction fidelity through alignment with vision foundation models and enhanced perceptual supervision.

In our framework, the low-frequency representation \( \mathbf{x}_L \) obtained from the wavelet decomposition is encoded into a latent embedding \( \mathbf{z}_L \), which is reconstructed through the decoder while being optimized under a combination of complementary objectives. Specifically, the training objective consists of the standard VAE reconstruction loss \( \mathcal{L}_{\text{rec}}^{L} \) and KL regularization \( \mathcal{L}_{\text{KL}}^{L} \), together with additional supervision terms introduced in VAVAE~\cite{yao2025vavae}, including a vision foundation alignment loss \( \mathcal{L}_{\text{VF}}^{L} \), an adversarial regularization objective \( \mathcal{L}_{\text{GAN}}^{L} \) inspired by VQGAN~\cite{esser2021taming}, and a perceptual similarity loss \( \mathcal{L}_{\text{LPIPS}}^{L} \)~\cite{zhang2018unreasonable}. The complete low-frequency objective is given by $\mathcal{L}^{L}$.
\[
\mathcal{L}^{L} = 
\mathcal{L}_{\text{rec}}^{L}
+ \beta\, \mathcal{L}_{\text{KL}}^{L}
+ \lambda_{\text{VF}}\, \mathcal{L}_{\text{VF}}^{L}
+ \lambda_{\text{GAN}}\, \mathcal{L}_{\text{GAN}}^{L}
+ \lambda_{\text{LPIPS}}\, \mathcal{L}_{\text{LPIPS}}^{L}.
\]
where,
\begin{align*}
\mathcal{L}_{\text{rec}}^{L} &= 
\mathbb{E}_{q(\mathbf{z}_L \mid \mathbf{x}_L)}
\left[ \left\| \mathbf{x}_L - D_L(\mathbf{z}_L) \right\|_2^2 \right], \\
\mathcal{L}_{\text{KL}}^{L} &= 
D_{\mathrm{KL}}\!\left(q(\mathbf{z}_L \mid \mathbf{x}_L)\,\|\,p(\mathbf{z}_L)\right),
\end{align*}
where \( \mathcal{L}_{\text{rec}}^{L} \) is the low-frequency reconstruction loss,  
\( \mathcal{L}_{\text{KL}}^{L} \) is the KL divergence between the encoder's approximate posterior \( q(\mathbf{z}_L \mid \mathbf{x}_L) \) and the Gaussian prior \( p(\mathbf{z}_L) = \mathcal{N}(\mathbf{0}, \mathbf{I}) \). The vision foundation loss \( \mathcal{L}_{\text{VF}}^L \) aligns the latent codes with the feature space of a pretrained foundation model (DINOv2~\cite{oquab2023dinov2}). The adversarial loss \( \mathcal{L}_{\text{GAN}}^L \) introduces a discriminator to distinguish real from reconstructed low-frequency inputs. Finally, the perceptual loss \( \mathcal{L}_{\text{LPIPS}}^L \) encourages perceptual similarity between input and reconstruction using features from a pretrained Inception network~\cite{szegedy2016rethinking}. All these loss components are commonly used in modern VAE frameworks. This overall objective encourages the latent representation to preserve global structure through reconstruction and KL regularization, while perceptual and adversarial supervision with vision foundation alignment loss improve visual fidelity and semantic alignment of the reconstructed images.

\vspace{0.2cm}

\noindent\textbf{High-Frequency Objective.} In contrast to low frequency optimization, high frequency representations are optimized without any supervision from pretrained models, as these pretrained models tend to be trained jointly with low-and high spatial frequencies. To mitigate any external biases, we simplify our training objective focusing on reconstruction objective along with KL and adversarial regularization. The overall high-frequency objective is given by $\mathcal{L}^{H}$.
\begin{align*}
\mathcal{L}^{H} &= \mathcal{L}_{\text{rec}}^H + \beta \cdot \mathcal{L}_{\text{KL}}^H + \mathcal{L}_{\text{GAN}}^H, \\
\text{where,} \\
\mathcal{L}_{\text{rec}}^H &= \mathbb{E}_{q(\mathbf{z}_H|\mathbf{x}_H)} \left[ \left\| \mathbf{x}_H - D_H(\mathbf{z}_H) \right\|_2 \right], \\
\mathcal{L}_{\text{KL}}^H &= D_{\mathrm{KL}}\left(q(\mathbf{z}_H \mid \mathbf{x}_H) \,\|\, p(\mathbf{z}_H)\right)
\end{align*}
\( \mathcal{L}_{\text{rec}}^{H} \) is the high-frequency reconstruction loss,  
\( \mathcal{L}_{\text{KL}}^{H} \) is the KL divergence between the encoder's approximate posterior \( q(\mathbf{z}_H \mid \mathbf{x}_H) \) and the Gaussian prior \( p(\mathbf{z}_H) = \mathcal{N}(\mathbf{0}, \mathbf{I}) \).

\vspace{0.2cm}

\noindent\textbf{Inference.} At inference time, both decoders $(D_{L}, D_{H})$ are used to reconstruct the respective frequency bands, and the final image is synthesized via inverse wavelet transform:
\[
\hat{\mathbf{x}}_L = D_L(\mathbf{z}_L), \quad \hat{\mathbf{x}}_H = D_H(\mathbf{z}_H), \quad
\hat{\mathbf{x}} = \mathcal{W}^{-1}(\hat{\mathbf{x}}_L, \hat{\mathbf{x}}_H).
\]
%
%\noindent
This frequency-aware latent representation learning via DeBaT yields expressive latent embeddings by preserving both coarse structure and fine details across the full spectrum of spatial frequencies.

\subsection{Latent Fusion for Generative Modeling}

To enable frequency-aware generation, we integrate the proposed DeBaT tokenizer into the latent diffusion framework LightningDiT~\cite{yao2025vavae}. Given an input image \( \mathbf{x} \), wavelet decomposition produces low- and high-frequency components \( (\mathbf{x}_L, \mathbf{x}_H) \), which are encoded into frequency-specific latent embeddings. 
\[ \mathbf{z}_L = E_L(\mathbf{x}_L), \qquad \mathbf{z}_H = E_H(\mathbf{x}_H), \] 
where \( \mathbf{z}_L \in \mathbb{R}^{H \times W \times C_L} \) captures global structure and \( \mathbf{z}_H \in \mathbb{R}^{H \times W \times C_H} \) encodes fine details. The two embeddings are fused through channel-wise concatenation yielding a unified latent representation \( \tilde{\mathbf{z}} \in \mathbb{R}^{H \times W \times (C_L + C_H)} \). The diffusion model then learns the latent distribution \( p_\theta(\tilde{\mathbf{z}}) \). During generation, the sampled latent \( \tilde{\mathbf{z}} \) is separated into respective low and high frequency embeddings. 
\[ \hat{\mathbf{z}}_L = \tilde{\mathbf{z}}_{1:C_L}, \qquad \hat{\mathbf{z}}_H = \tilde{\mathbf{z}}_{C_L+1:C_L+C_H}, \] 

which are decoded independently $\hat{\mathbf{x}}_L = D_L(\hat{\mathbf{z}}_L)$, and $\hat{\mathbf{x}}_H = D_H(\hat{\mathbf{z}}_H),$ 
and the image is reconstructed via inverse wavelet transform 
$\hat{\mathbf{x}} = \mathcal{W}^{-1}(\hat{\mathbf{x}}_L, \hat{\mathbf{x}}_H).$ This enables the diffusion model to operate on a frequency-aware latent space while preserving modularity with existing latent generative pipelines.

\section{Experiments}

To evaluate DeBaT, we first access the reconstruction performance of visual tokenizers in both spatial and spectral domains. Then, we integrate the latent embeddings into standard image sythesis pipeline, comparing DeBaT combined with LightningDiT~\cite{yao2025vavae} to competing models. All experiments are conducted on ImageNet-1K~\cite{deng2009imagenet} and use 50k ImageNet validation samples for evaluation. Further architectural and implementation details of DeBaT are provided in Appendix section A \& section B.

%~\autoref{supp-sec:debat-architecture} \& ~\autoref{supp-sec:implementation}. 

\subsection{Performance Evaluation of Latent Tokenizers} 
To evaluate the reconstruction performance of latent embeddings in both the spatial and spectral (frequency) domains, we consider visual tokenizers commonly used in modern latent generative models. These tokenizers are different variants of variational autoencoding architectures, each aiming to learn compact and informative latent embedding representations, which depict the underlying input data distribution. For comprehensive analysis, we include recent discrete~\cite{yu2024image, esser2021taming, tian2024visual} and continuous latent tokenizers~\cite{chen2024deep, rombach2022high, li2024autoregressive, lee2022autoregressive, yu2024image, yao2025vavae}. We compare these latent tokenizers  with DeBaT. This enables comparison across various latent tokenizers with varying latent dimensionalities and feature resolutions as shown in ~\autoref{tab:reconstruction_detailed}. We use the metrics discussed in ~\autoref{table1_metrics} for evaluation.

%%%%%%%%%%%%%%%%%%%%%%%%%%%%%%%%%%%%%%%%%%%%%%%%%%%%%%%%%%%%%%%%%%%%%%%%%%%%%%%%%%%%%%%

\begin{table*}[t]
\centering
\resizebox{\linewidth}{!}{
\scriptsize
\setlength{\tabcolsep}{6pt}
\renewcommand{\arraystretch}{0.9}
\rowcolors{2}{gray!10}{white}
\begin{tabular}{llccccc}
\toprule
\textbf{Visual Tokenizer} & \textbf{Tokenizer Latent} & \textbf{Rec.} & \textbf{LF} & \textbf{HF} & \textbf{LPIPS} & \textbf{rFID} \\
\midrule
DC-AE~\cite{chen2024deep} & 8x8x32 & 0.0194 & 0.0484 & 0.0097 & 0.1580 & 0.7450 \\
DC-AE~\cite{chen2024deep} & 4x4x128 & 0.0207 & 0.0527 & 0.0100 & 0.1667 & 0.7623 \\
DC-AE~\cite{chen2024deep} & 4×4×512 & 0.0225 & 0.0581 & 0.0106 & 0.1805 & 0.7912 \\
SD-VAE~\cite{rombach2022high} & 16x16x16 & 0.0180 & 0.0422 & 0.0100 & 0.1743 & 0.6213 \\
KL-VAE~\cite{li2024autoregressive} & 16x16x16 & 0.0148 & 0.0326 & 0.0089 & 0.1355 & 0.5318 \\
MS-VQ-VAE~\cite{tian2024visual} &  16x16x32 (4096) & 0.0195 & 0.0549 & 0.0076 & 0.1890 & 0.6981 \\
RQ-VAE~\cite{lee2022autoregressive} & 8x8x16 (16384) & 0.0416 & 0.1219 & 0.0149 & 0.2712 & 0.9095 \\
TiTok-VQ-VAE~\cite{yu2024image} & 128x64 (8192) & 0.0226 & 0.0561 & 0.0114 & 0.2082 & 0.7550 \\  %512f256x2c64v8192 
TiTok-VQ-VAE~\cite{yu2024image} & 64x64 (8192) & 0.0339 & 0.0947 & 0.0137 & 0.2657 & 0.8823 \\
TiTok-VQ-VAE~\cite{yu2024image} & 32x64 (8192) & 0.0450 & 0.1344 & 0.0152 & 0.2949 & 0.9416 \\
TiTok-VAE~\cite{yu2024image} & 256x16 & 0.0332 & 0.0923 & 0.0134 & 0.2232 & 0.8640 \\
TiTok-VAE~\cite{yu2024image} & 128x16 & 0.0461 & 0.1380 & 0.0155 & 0.2682 & 0.9187 \\
TiTok-VAE~\cite{yu2024image} & 64x16 & 0.0617 & 0.1961 & 0.0170 & 0.3182 & 0.9761 \\
VQ-VAE~\cite{esser2021taming} & 16x16x256 (1024) & 0.0492 & 0.1478 & 0.0163 & 0.3064 & 0.9102 \\
VQ-VAE~\cite{esser2021taming} & 16x16x256 (16384) & 0.0438 & 0.1262 & 0.0164 & 0.2784 & 0.8826 \\
\midrule
LiteVAE~\cite{sadat2024litevae} & 32x32x12 & 0.0086 & 0.0179 & 0.0050 & 0.0970 & 0.2250 \\

VA-VAE~\cite{yao2025vavae} & 16x16x32 & 0.0105 & 0.0200 & 0.0074 & 0.0975 & 0.2777 \\

\textbf{DeBaT (Ours)} & \textbf{16x16x32} & \textbf{0.0044} & \textbf{0.0114} & \textbf{0.0020} & \textbf{0.0940} & \textbf{0.2156} \\
\bottomrule
\end{tabular}
}
\vspace{1pt}
\caption{ Quantitative comparison of various latent tokenizers based on their reconstruction quality across multiple metrics. Tokenizer latent represents latent dimensionality (height $\times$ width $\times $ channel), in case of quantized latents  latent dimensionality is represented by (height $\times$ width $\times $ channel (codebook size)). For TikTok versions as the latent space is 1-D, latent dimension represents (number of tokens $\times $ channels). Metrics: \textbf{Rec.} (reconstruction loss), \textbf{LF} (low-frequency loss), \textbf{HF} (high-frequency loss), \textbf{LPIPS} and \textbf{rFID}. Lower is better. DeBaT achieves a substantial improvement over previous methods in terms of reconstruction quality.}
\label{tab:reconstruction_detailed}
\end{table*}

%%%%%%%%%%%%%%%%%%%%%%%%%%%%%%%%%%%%%%%%%%%%%%%%%%%%%%%%%%%%%%%%%%%%%%%%%%%%%%%%%%%%%%

\begin{figure*}[th!]
\centering\vspace{-2mm}
\begin{subfigure}[b]{0.135\textwidth}
    \centering
    \textbf{\scriptsize Input}\\[2pt]
    \includegraphics[width=\linewidth]{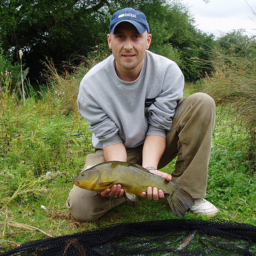}
    \caption*{\scriptsize Recon. Error:}
\end{subfigure}
\hfill
\begin{subfigure}[b]{0.135\textwidth}
    \centering
    \textbf{\scriptsize VA-VAE}\\[2pt]
    \includegraphics[width=\linewidth]{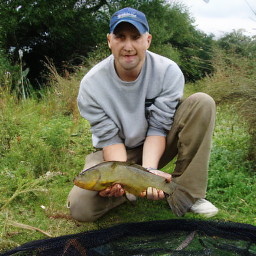}
    \caption*{\scriptsize 0.0044}
\end{subfigure}
\hfill
\begin{subfigure}[b]{0.135\textwidth}
    \centering
    \textbf{\scriptsize Lite-VAE}\\[2pt]
    \includegraphics[width=\linewidth]{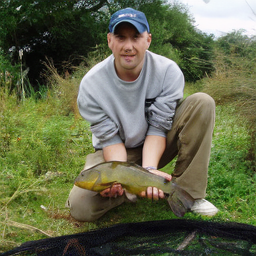}
    \caption*{\scriptsize 0.0028}
\end{subfigure}
\hfill
\begin{subfigure}[b]{0.135\textwidth}
    \centering
    \textbf{\scriptsize Ours }\\[2pt]
    \includegraphics[width=\linewidth]{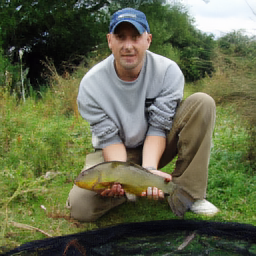}
    \caption*{\scriptsize 0.0018}
\end{subfigure}
\hfill
\begin{subfigure}[b]{0.135\textwidth}
    \centering
    \textbf{\scriptsize VA-VAE}\\[2pt]
    \includegraphics[width=\linewidth]{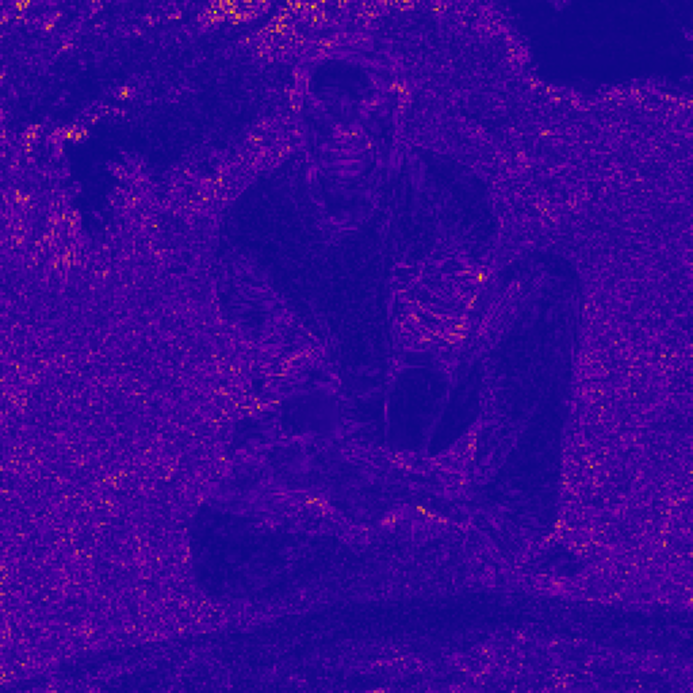}
    \caption*{\scriptsize Error Map}
\end{subfigure}
\hfill
\begin{subfigure}[b]{0.135\textwidth}
    \centering
    \textbf{\scriptsize Lite-VAE}\\[2pt]
    \includegraphics[width=\linewidth]{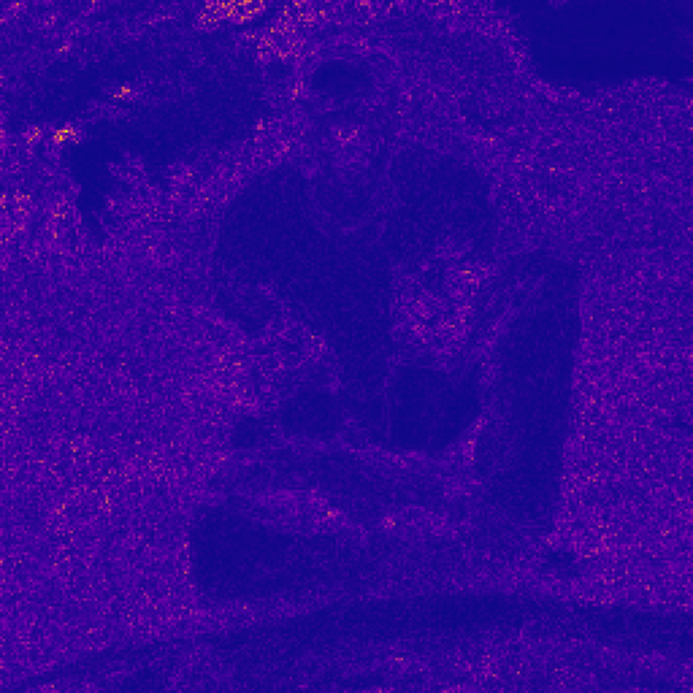}
    \caption*{\scriptsize Error Map}
\end{subfigure}
\hfill
\begin{subfigure}[b]{0.135\textwidth}
    \centering
    \textbf{\scriptsize Ours}\\[2pt]
    \includegraphics[width=\linewidth]{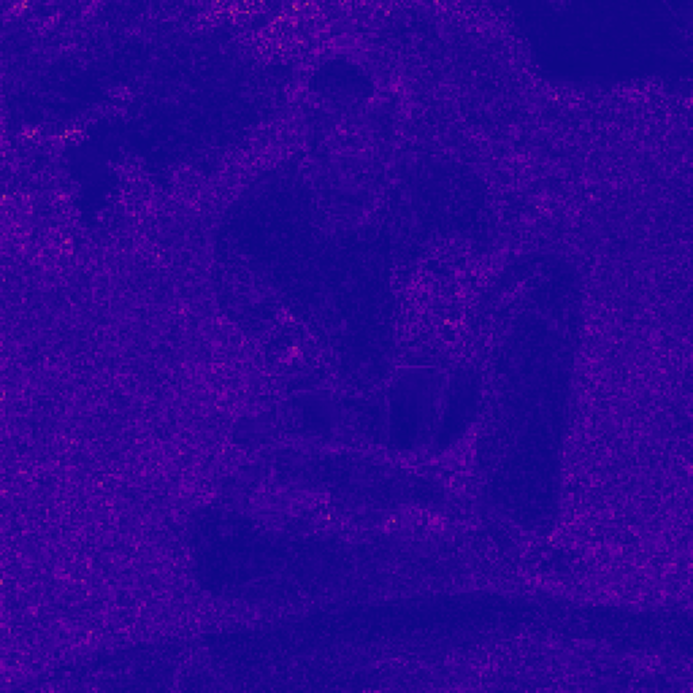}
    \caption*{\scriptsize Error Map}
\end{subfigure}

%%%%%%%%%%%%%%%%%%%%%%%%%%%%%%%%%%%%%%%%%%%%%%%%%%%%%%%%%%%%%

\begin{subfigure}[b]{0.135\textwidth}
    \centering
   % \textbf{Input}\\[2pt]
    \includegraphics[width=\linewidth]{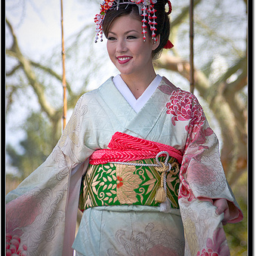}
    \caption*{\scriptsize Recon. Error:}
\end{subfigure}
\hfill
\begin{subfigure}[b]{0.135\textwidth}
    \centering
  %  \textbf{VA-VAE}\\[2pt]
    \includegraphics[width=\linewidth]{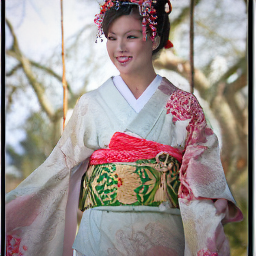}
    \caption*{\scriptsize 0.0040}
\end{subfigure}
\hfill
\begin{subfigure}[b]{0.135\textwidth}
    \centering
   % \textbf{WF-VAE (Ours)}\\[2pt]
    \includegraphics[width=\linewidth]{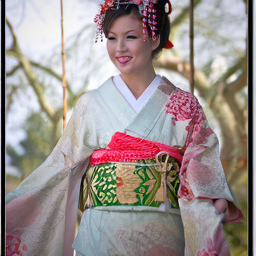}
    \caption*{\scriptsize 0.0023}
\end{subfigure}
\hfill
\begin{subfigure}[b]{0.135\textwidth}
    \centering
  %  \textbf{VA-VAE Error Map}\\[2pt]
    \includegraphics[width=\linewidth]{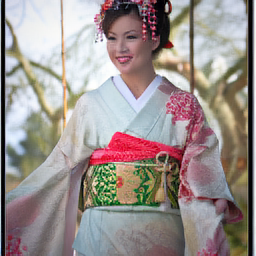}
    \caption*{\scriptsize 0.0018}
\end{subfigure}
\hfill
\begin{subfigure}[b]{0.135\textwidth}
    \centering
  %  \textbf{VA-VAE Error Map}\\[2pt]
    \includegraphics[width=\linewidth]{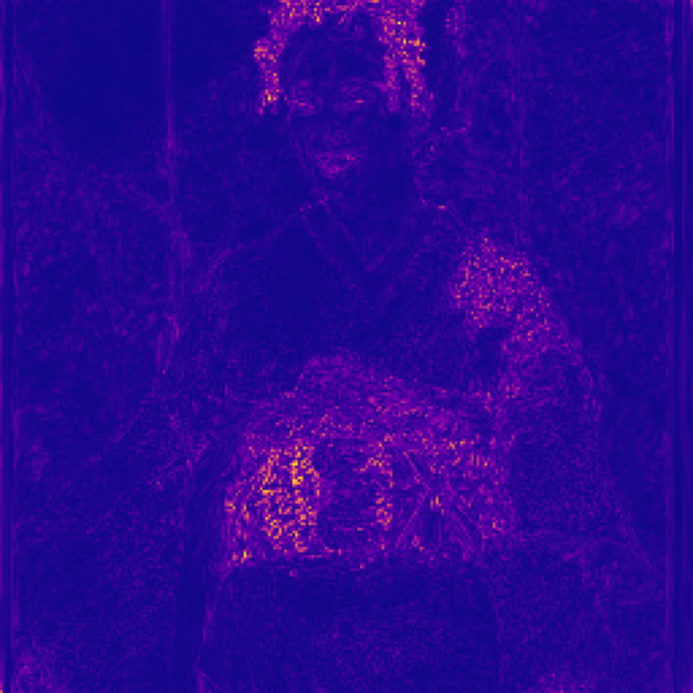}
    \caption*{\scriptsize Error Map}
\end{subfigure}
\hfill
\begin{subfigure}[b]{0.135\textwidth}
    \centering
  %  \textbf{VA-VAE Error Map}\\[2pt]
    \includegraphics[width=\linewidth]{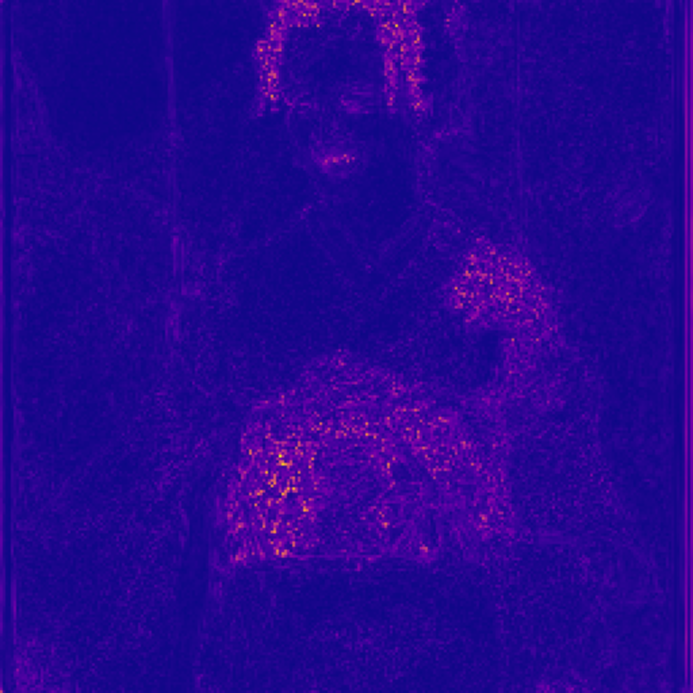}
    \caption*{\scriptsize Error Map}
\end{subfigure}
\hfill
\begin{subfigure}[b]{0.135\textwidth}
    \centering
   % \textbf{WF-VAE Error Map}\\[2pt]
    \includegraphics[width=\linewidth]{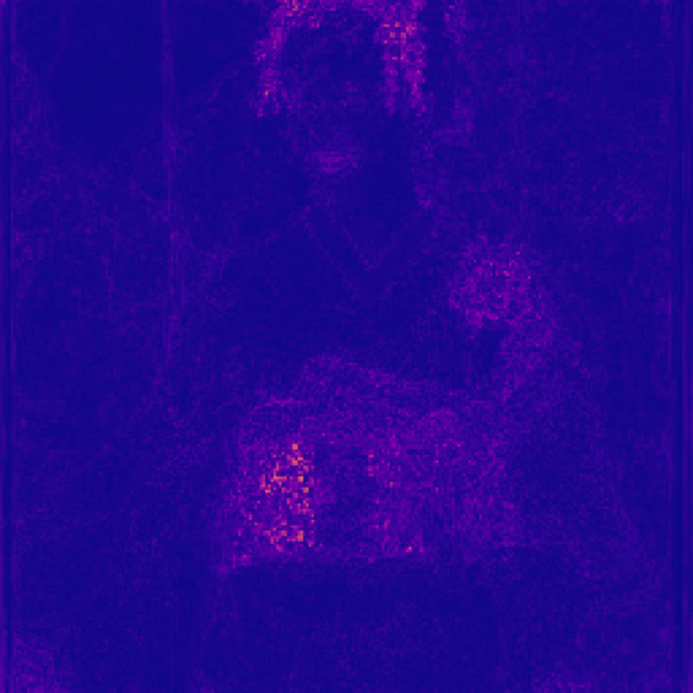}
    \caption*{\scriptsize Error Map}
\end{subfigure}

% ==========================================
% ======= Example 3 (00003) ================
% ==========================================
\begin{subfigure}[b]{0.135\textwidth}
    \centering
   % \textbf{Input}\\[2pt]
    \includegraphics[width=\linewidth]{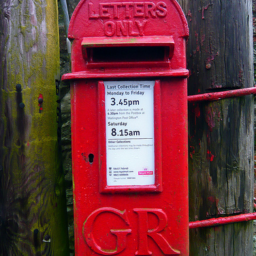}
    \caption*{\scriptsize Recon. Error:}
\end{subfigure}
\hfill
\begin{subfigure}[b]{0.135\textwidth}
    \centering
  %  \textbf{VA-VAE}\\[2pt]
    \includegraphics[width=\linewidth]{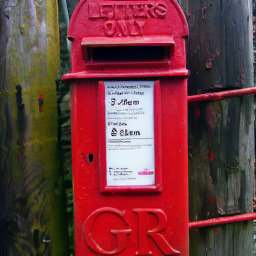}
    \caption*{\scriptsize 0.0025}
\end{subfigure}
\hfill
\begin{subfigure}[b]{0.135\textwidth}
    \centering
   % \textbf{WF-VAE (Ours)}\\[2pt]
    \includegraphics[width=\linewidth]{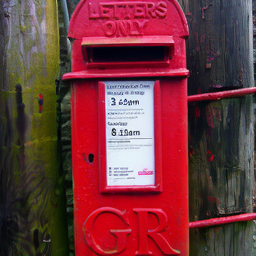}
    \caption*{\scriptsize 0.0015}
\end{subfigure}
\hfill
\begin{subfigure}[b]{0.135\textwidth}
    \centering
  %  \textbf{VA-VAE Error Map}\\[2pt]
    \includegraphics[width=\linewidth]{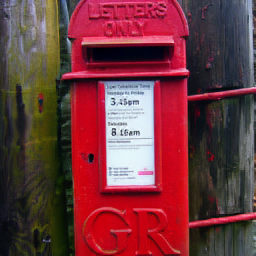}
    \caption*{\scriptsize 0.0012}
\end{subfigure}
\hfill
\begin{subfigure}[b]{0.135\textwidth}
    \centering
  %  \textbf{VA-VAE Error Map}\\[2pt]
    \includegraphics[width=\linewidth]{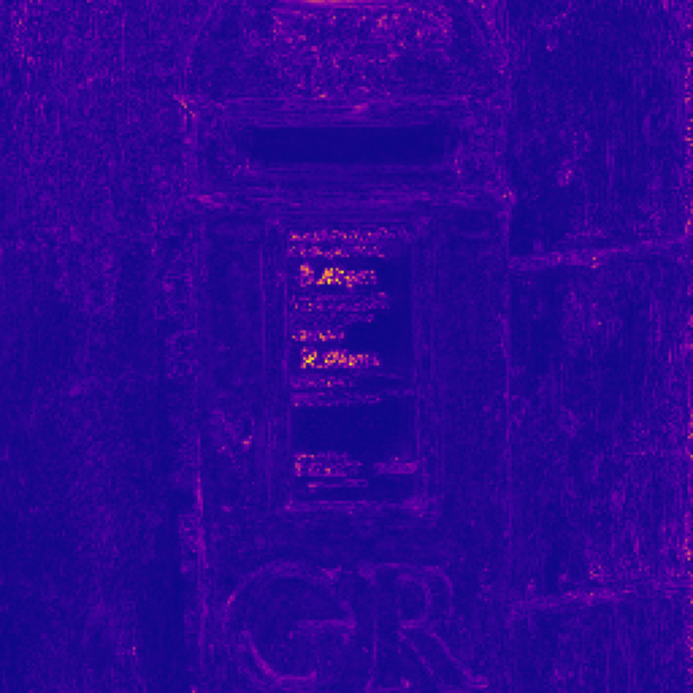}
    \caption*{\scriptsize Error Map}
\end{subfigure}
\hfill
\begin{subfigure}[b]{0.135\textwidth}
    \centering
  %  \textbf{VA-VAE Error Map}\\[2pt]
    \includegraphics[width=\linewidth]{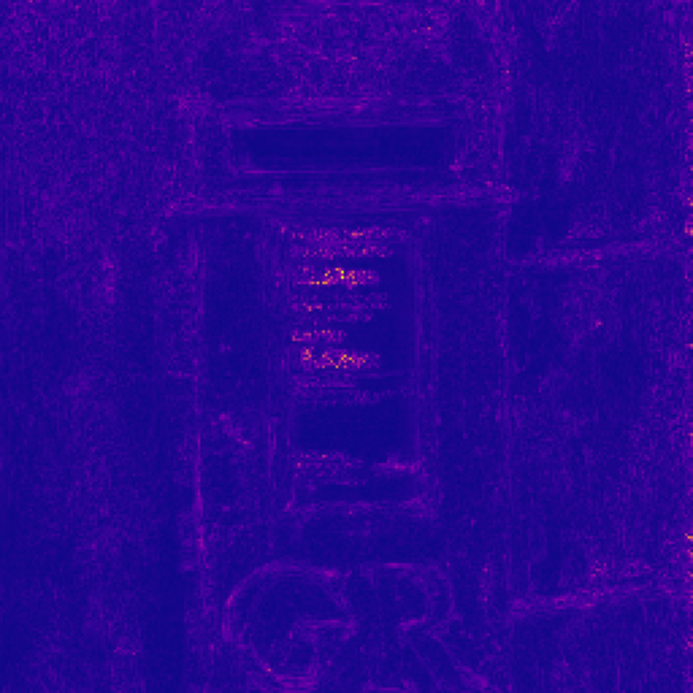}
    \caption*{\scriptsize Error Map}
\end{subfigure}
\hfill
\begin{subfigure}[b]{0.135\textwidth}
    \centering
   % \textbf{WF-VAE Error Map}\\[2pt]
    \includegraphics[width=\linewidth]{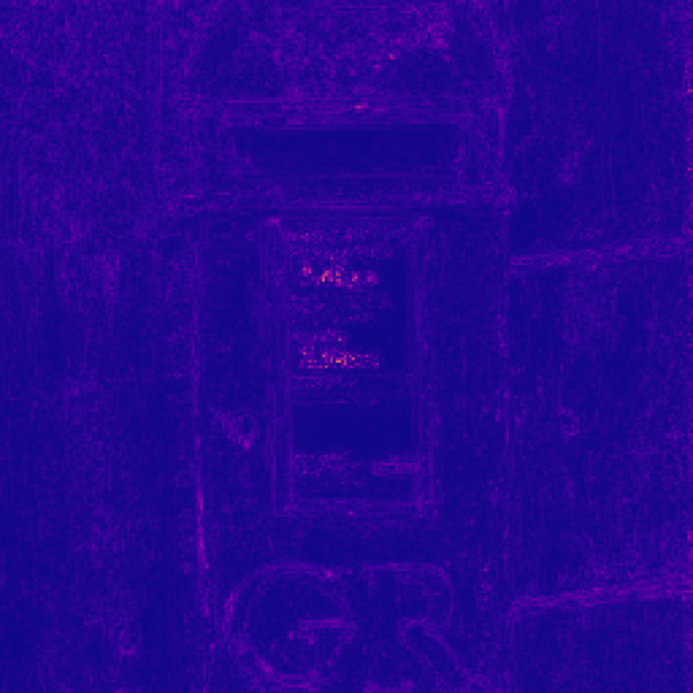}
    \caption*{\scriptsize Error Map}
\end{subfigure}
\vspace{-2mm}
\caption{Comparison with respect to reconstructions and error maps with best baseline VA-VAE and the previous wavelet base method LiteVAE. DeBaT Error Maps have consistently lower magnitudes than VA-VAE and LiteVAE Error Maps, visible as fine details that are correctly reconstructed in particular in structured regions like person faces, the texture of the dress or the printed text. LiteVAE performs better than VA-VAE but makes more errors in fine details than DeBaT.}
\label{fig:examples_fiveviews}
\end{figure*}

Our analysis in ~\autoref{tab:reconstruction_detailed} reveals that KL-regularized VAE from~\cite{li2024autoregressive} with right parameterizations consistently outperform both traditional VAE and vector quantization (VQ-VAE) baselines across all reconstruction metrics. Also KL-regularized VAE show efficient reconstruction performance at much better compression rate with lower dimensionality. This improvement stems from the absence of quantization artifacts, which often degrade reconstruction quality in discrete latent spaces. Additionally, the effectiveness of KL-VAE is closely tied to the quality of their latent parameterization schemes. Notably, KL-VAE employed in recent autoregressive models~\cite{chen2025masked}. VA-VAE~\cite{yao2025vavae} further aligns its latent space using KL-VAE parameterization with foundation models like DINOv2~\cite{oquab2023dinov2} achieving strong performance across low and high-frequency reconstruction, along with improved perceptual quality. However, these models still rely on a joint optimization of spatial frequencies, which can lead to trade-offs and suboptimal detail preservation, especially in the high-frequency spectrum. In contrast, our proposed approach DeBaT builds on the VA-VAE framework by explicitly decoupling and modeling low- and high-frequency components in learning of the compact and frequency preserving expressive latent embeddings. This separation enables focused representation learning for each frequency band, leading to more precise reconstructions of both global structures and fine details. For additional quantitative evaluation results, please refer to Table 1 and Table 2 in the Appendix section C. ~\autoref{fig:examples_fiveviews} shows visual comparison with best baselines VAVAE, LiteVAE and our approach with reconstruction errors and error maps, where our reconstructions are much detailed with lower reconstruction error and lower magnitude in error maps overall, showcasing preservation of high frequency details. In comparison to previous work leveraging wavelet transforms~\cite{sadat2024litevae}, we observe that DeBaT achieves a substantially lower reconstruction loss, particularly prominent in the high frequency loss. Notably, LiteVAE employs wavelets to facilitate highly compressed latent codes, not fully leveraging their ability to decouple frequency bands as proposed in DeBaT. %Therefore, while DeBaT allows for more faithful latent encoding, the two approaches can not be directly compared and have to be seen as complementary in practice. 

\begin{table*}[t]
\centering
\begingroup
\scriptsize
\setlength{\tabcolsep}{2.6pt}
\renewcommand{\arraystretch}{0.92}
\rowcolors{2}{gray!10}{white}
\resizebox{\linewidth}{!}{
\begin{tabular}{l@{\hspace{0mm}}c@{\hspace{0.6mm}}c@{\hspace{0.5mm}}c|ccccc|ccccc}
\toprule
\multirow{2}{*}{\textbf{Method}} &
\multirow{2}{*}{\textbf{Tokenizer}} &
\multirow{2}{*}{\textbf{Epochs}} &
\multirow{2}{*}{\textbf{\#Params}} &
\multicolumn{5}{c|}{\textbf{Generation w/o CFG}} &
\multicolumn{5}{c}{\textbf{Generation w/ CFG}} \\
\cmidrule(lr){5-9}\cmidrule(lr){10-14}
& & & &
\textbf{gFID} & \textbf{sFID} & \textbf{IS} & \textbf{Pre.} & \textbf{Rec.} &
\textbf{gFID} & \textbf{sFID} & \textbf{IS} & \textbf{Pre.} & \textbf{Rec.} \\
\midrule

\multicolumn{14}{c}{\textit{Autoregressive (AR)}} \\
MaskGIT~\cite{maskgit}   & VQ-VAE     & 555  & 227M & 6.18 & -    & 182.1 & 0.80 & 0.51 & -    & -    & -    & -    & -    \\
LlamaGen~\cite{llamagen}  & VQ-VAE     & 300  & 3.1B & 9.38 & 8.24 & 112.9 & 0.69 & 0.67 & 2.18 & 5.97 & 263.3 & 0.81 & 0.58 \\
VAR~\cite{tian2024visual}       & MS-VQ-VAE  & 350  & 2.0B & -    & -    & -     & -    & -    & 1.80 & -    & 365.4 & 0.83 & 0.57 \\
MagViT-v2~\cite{magvitv2} & VQ-VAE     & 1080 & 307M & 3.65 & -    & 200.5 & -    & -    & 1.78 & -    & 319.4 & -    & -    \\
MAR~\cite{li2024autoregressive}       & KL-VAE     & 800  & 945M & 2.35 & -    & 227.8 & 0.79 & 0.62 & 1.55 & -    & 303.7 & 0.81 & 0.62 \\

\midrule
\multicolumn{14}{c}{\textit{Latent Diffusion Models}} \\
MaskDiT~\cite{maskdit}   & SD-VAE & 1600 & 675M & 5.69 & 10.34 & 177.9 & 0.74 & 0.60 & 2.28 & 5.67 & 276.6 & 0.80 & 0.61 \\
DiT~\cite{dit}       & SD-VAE & 1400 & 675M & 9.62 & 6.85  & 121.5 & 0.67 & 0.67 & 2.27 & 4.60 & 278.2 & 0.83 & 0.57 \\
SiT~\cite{sit}       & SD-VAE & 1400 & 675M & 8.61 & 6.32  & 131.7 & 0.68 & 0.67 & 2.06 & 4.50 & 270.3 & 0.82 & 0.59 \\
FasterDiT~\cite{fasterdit} & SD-VAE & 400  & 675M & 7.91 & 5.45  & 131.3 & 0.67 & 0.69 & 2.03 & 4.63 & 264.0 & 0.81 & 0.60 \\
MDT~\cite{mdt}       & SD-VAE & 1300 & 675M & 6.23 & 5.23  & 143.0 & 0.71 & 0.65 & 1.79 & 4.57 & 283.0 & 0.81 & 0.61 \\
MDTv2~\cite{mdtv2}     & SD-VAE & 1080 & 675M & -    & -     & -     & -    & -    & 1.58 & 4.52 & 314.7 & 0.79 & 0.65 \\
REPA~\cite{repa}      & SD-VAE & 800  & 675M & 5.90 & -     & -     & -    & -    & 1.42 & 4.70 & 305.7 & 0.80 & 0.65 \\

\midrule

LightningDiT~\cite{yao2025vavae} & Lite-VAE & 64 & 675M &
7.97 & 5.13 & 128.1 & 0.70 & 0.64 &
2.27 & 4.50 & 260.0 & 0.80 & 0.57 \\

LightningDiT~\cite{yao2025vavae} & VA-VAE & 64 & 675M &
5.14 & 4.22 & 130.2 & 0.76 & 0.62 &
2.11 & 4.16 & 252.3 & 0.81 & 0.58 \\

LightningDiT~\cite{yao2025vavae} & \textbf{DeBaT} & 64 & 675M &
\textbf{3.24} & \textbf{4.09} & \textbf{193.7} & \textbf{0.83} & \textbf{0.69} &
\textbf{1.52} & \textbf{4.07} & \textbf{317.4} & \textbf{0.83} & \textbf{0.65} \\

\bottomrule

\end{tabular}
}
\vspace{1pt}
\caption{Comparision of generation performance of autoregressive and latent diffusion models with and without classifier-free guidance (CFG) vs.~LightningDiT with our DeBaT tokenizer. Missing values are indicated by \texttt{-}. We report metrics available in the respective works. DeBaT provides consistent improvement over previous works, in particular over VA-VAE where the direct comparison in terms of generation method (LightningDiT), epochs and number of parameters is straight-forward.}
\label{tab:generation_metrics}
\endgroup
\end{table*}

\begin{figure*}[t]
\centering

%%%%%%%%%%%%%%%%%%%%%%%%%%%%%%%%%%%%%%%%%%%%%%%%%%%%
%%%%%%%%%%%%%%%% Example 1 %%%%%%%%%%%%%%%%%%%%%%%%%
%%%%%%%%%%%%%%%%%%%%%%%%%%%%%%%%%%%%%%%%%%%%%%%%%%%%

\begin{subfigure}[b]{0.16\textwidth}
\centering
%\textbf{\scriptsize Generation}\\[2pt]
\includegraphics[width=\linewidth]{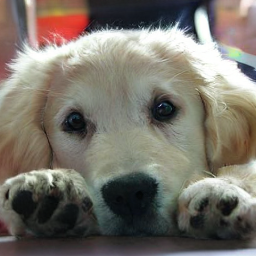}
%\caption*{\scriptsize L2 Error:}
\end{subfigure}
\hfill
\begin{subfigure}[b]{0.16\textwidth}
\centering
%\textbf{\scriptsize \scriptsize Train Image}\\[2pt]
\includegraphics[width=\linewidth]{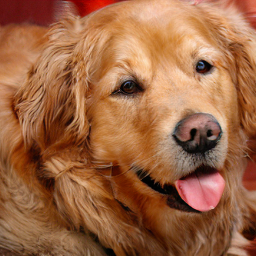}
%\caption*{\scriptsize 0.2117}
\end{subfigure}
\hfill
\begin{subfigure}[b]{0.16\textwidth}
\centering
%\textbf{\scriptsize Generation}\\[2pt]
\includegraphics[width=\linewidth]{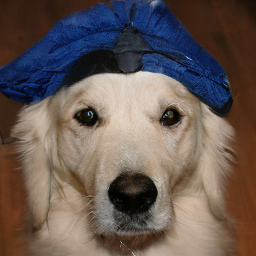}
%\caption*{\scriptsize L2 Error:}
\end{subfigure}
\hfill
\begin{subfigure}[b]{0.16\textwidth}
\centering
%\textbf{\scriptsize Train Image}\\[2pt]
\includegraphics[width=\linewidth]{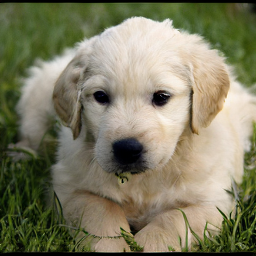}
%\caption*{\scriptsize 0.1497}
\end{subfigure}
\hfill
\begin{subfigure}[b]{0.16\textwidth}
\centering
%\textbf{\scriptsize Generation}\\[2pt]
\includegraphics[width=\linewidth]{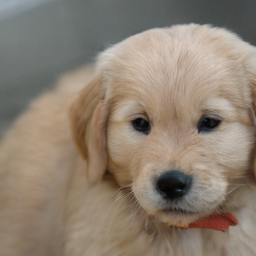}
%\caption*{\scriptsize L2 Error:}
\end{subfigure}
\hfill
\begin{subfigure}[b]{0.16\textwidth}
\centering
%\textbf{\scriptsize Train Image}\\[2pt]
\includegraphics[width=\linewidth]{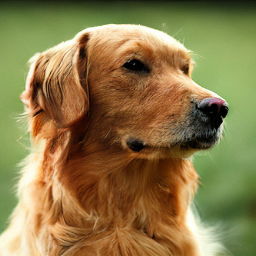}
%\caption*{\scriptsize 0.2415}
\end{subfigure}

%%%%%%%%%%%%%%%%%%%%%%%%%%%%%%%%%%%%%%%%%%%%%%%%%%%%
%%%%%%%%%%%%%%%% Example 2 %%%%%%%%%%%%%%%%%%%%%%%%%
%%%%%%%%%%%%%%%%%%%%%%%%%%%%%%%%%%%%%%%%%%%%%%%%%%%%

\begin{subfigure}[b]{0.16\textwidth}
\centering
\includegraphics[width=\linewidth]{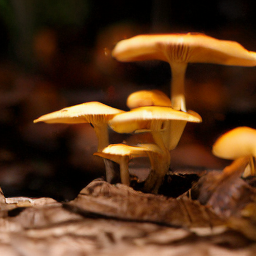}
%\caption*{\scriptsize L2 Error:}
\end{subfigure}
\hfill
\begin{subfigure}[b]{0.16\textwidth}
\centering
\includegraphics[width=\linewidth]{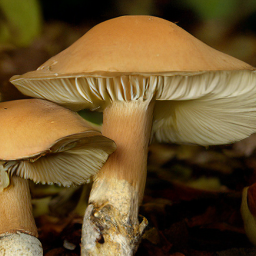}
%\caption*{\scriptsize 0.2120}
\end{subfigure}
\hfill
\begin{subfigure}[b]{0.16\textwidth}
\centering
\includegraphics[width=\linewidth]{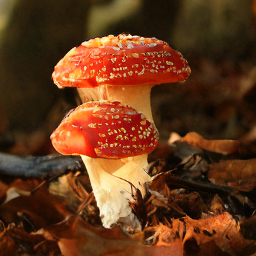}
%\caption*{\scriptsize L2 Error:}
\end{subfigure}
\hfill
\begin{subfigure}[b]{0.16\textwidth}
\centering
\includegraphics[width=\linewidth]{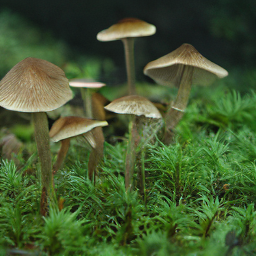}
%\caption*{\scriptsize 0.1973}
\end{subfigure}
\hfill
\begin{subfigure}[b]{0.16\textwidth}
\centering
\includegraphics[width=\linewidth]{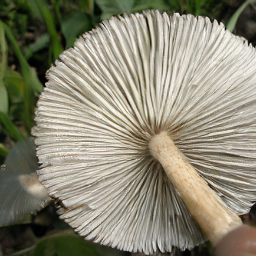}
%\caption*{\scriptsize L2 Error:}
\end{subfigure}
\hfill
\begin{subfigure}[b]{0.16\textwidth}
\centering
\includegraphics[width=\linewidth]{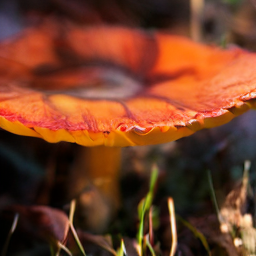}
%\caption*{\scriptsize 0.2319}
\end{subfigure}

%%%%%%%%%%%%%%%%%%%%%%%%%%%%%%%%%%%%%%%%%%%%%%%%%%%%
%%%%%%%%%%%%%%%% Example 3 %%%%%%%%%%%%%%%%%%%%%%%%%
%%%%%%%%%%%%%%%%%%%%%%%%%%%%%%%%%%%%%%%%%%%%%%%%%%%%

\begin{subfigure}[b]{0.16\textwidth}
\centering
\includegraphics[width=\linewidth]{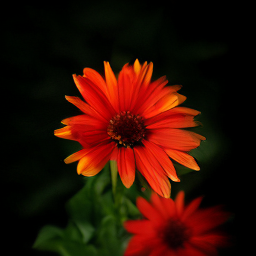}
%\caption*{\scriptsize L2 Error:}
\end{subfigure}
\hfill
\begin{subfigure}[b]{0.16\textwidth}
\centering
\includegraphics[width=\linewidth]{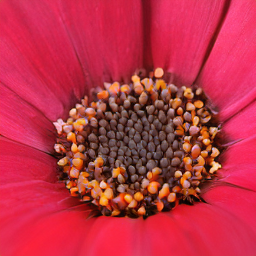}
%\caption*{\scriptsize 0.2526}
\end{subfigure}
\hfill
\begin{subfigure}[b]{0.16\textwidth}
\centering
\includegraphics[width=\linewidth]{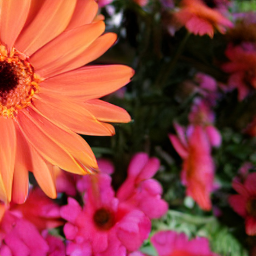}
%\caption*{\scriptsize L2 Error:}
\end{subfigure}
\hfill
\begin{subfigure}[b]{0.16\textwidth}
\centering
\includegraphics[width=\linewidth]{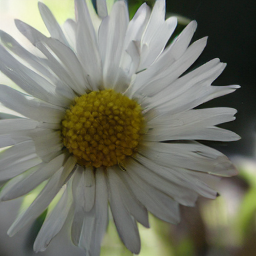}
%\caption*{\scriptsize 0.2325}
\end{subfigure}
\hfill
\begin{subfigure}[b]{0.16\textwidth}
\centering
\includegraphics[width=\linewidth]{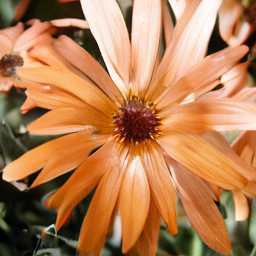}
%\caption*{\scriptsize L2 Error:}
\end{subfigure}
\hfill
\begin{subfigure}[b]{0.16\textwidth}
\centering
\includegraphics[width=\linewidth]{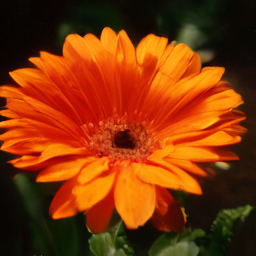}
%\caption*{\scriptsize 0.2558}
\end{subfigure}

%%%%%%%%%%%%%%%%%%%%%%%%%%%%%%%%%%%%%%%%%%%%%%%%%%%%
%%%%%%%%%%%%%%%% Example 4 %%%%%%%%%%%%%%%%%%%%%%%%%
%%%%%%%%%%%%%%%%%%%%%%%%%%%%%%%%%%%%%%%%%%%%%%%%%%%%

\begin{subfigure}[b]{0.16\textwidth}
\centering
\includegraphics[width=\linewidth]{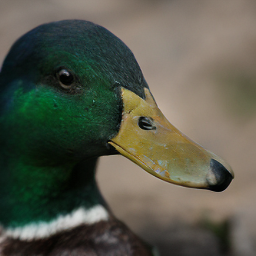}
%\caption*{\scriptsize L2 Error:}
\end{subfigure}
\hfill
\begin{subfigure}[b]{0.16\textwidth}
\centering
\includegraphics[width=\linewidth]{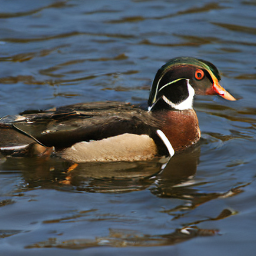}
%\caption*{\scriptsize 0.2526}
\end{subfigure}
\hfill
\begin{subfigure}[b]{0.16\textwidth}
\centering
\includegraphics[width=\linewidth]{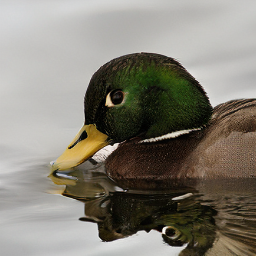}
%\caption*{\scriptsize L2 Error:}
\end{subfigure}
\hfill
\begin{subfigure}[b]{0.16\textwidth}
\centering
\includegraphics[width=\linewidth]{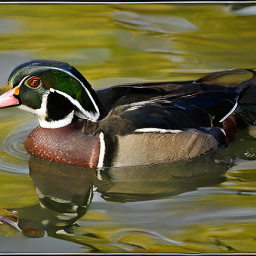}
%\caption*{\scriptsize 0.2325}
\end{subfigure}
\hfill
\begin{subfigure}[b]{0.16\textwidth}
\centering
\includegraphics[width=\linewidth]{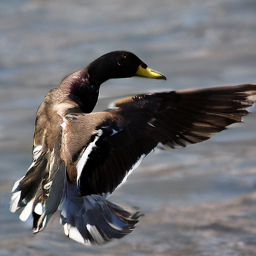}
%\caption*{\scriptsize L2 Error:}
\end{subfigure}
\hfill
\begin{subfigure}[b]{0.16\textwidth}
\centering
\includegraphics[width=\linewidth]{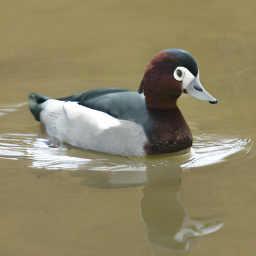}
%\caption*{\scriptsize 0.2558}
\end{subfigure}

\caption{Class-conditional generation results of classes : Dog (class id : 207), mushroom (class id : 947), Daisy (class id : 985), duck( class id : 97)
of LightningDiT with DeBaT.}
\label{fig:gen_main}

\end{figure*}

\subsection{Latent Generation Results}

We evaluate the generative quality of our learned latent embeddings by training a diffusion-based generative model on top of the DeBaT latent representations. Specifically, we use LightningDiT~\cite{yao2025vavae}, a fast-converging latent diffusion backbone to model the generation process, and compare against a wide range of state-of-the-art autoregressive and latent diffusion models, both with and without classifier-free guidance (CFG). Table~\ref{tab:generation_metrics} reports results across standard metrics: generation FID score (gFID), spatial FID (sFID), Inception Score (IS), Precision, and Recall. Lower gFID and sFID values indicate better image fidelity and diversity, while higher IS, precision, and recall reflect better sample quality. sFID is equivalent to FID but uses intermediate spatial features in the inception network rather than the spatially-pooled features in standard FID. 

\autoref{tab:generation_metrics} shows comparison of LightningDiT with DeBaT with other standard latent generation baselines. Our approach, LightningDiT with DeBaT, achieves the best gFID among all diffusion based generation methods, along with highly competitive sFID and strong scores across IS, precision, and recall with and without the CFG setting. Compared to auto-regressive models like MaskGIT~\cite{maskgit}, MAR~\cite{li2024autoregressive}, VAR~\cite{tian2024visual}, MagViT-v2~\cite{magvitv2} and LlamaGen~\cite{llamagen}, as well as latent diffusion models such as MaskDiT~\cite{maskdit}, DiT~\cite{dit}, SiT~\cite{sit}, FasterDiT~\cite{fasterdit}, MDF~\cite{mdt} and REPA~\cite{repa}, our method consistently delivers a better fidelity and diversity. Notably, it outperforms the LightningDiT baseline trained on VA-VAE embeddings across all metrics, demonstrating the advantage of our frequency-aware latent space in enhancing generation quality. These results validate the effectiveness of our DeBaT tokenizer for learning high-quality, detail-preserving representations that generalize well in generative pipelines. The qualitative results of our generated visualizations are shown in Figure~\ref{fig:gen_main}. Figure shows the conditional  generation samples of our approach when integrated with LightningDiT, showcasing diversity and quality in generated samples. Further quantitative evaluations with longer generative training are provided in Appendix subsection C.3.

\section{Ablation Study}

\noindent\textbf{Ablation on Frequency Decoupling.} To further investigate the contribution of decoupling low- and high-frequency components in our frequency-aware training strategy, we conduct an ablation study utilizing standard KL-VAE~\cite{li2024autoregressive}, multi-scale VQ-VAE~\cite{tian2024visual} and VAVAE~\cite{yao2025vavae} with coupled low and high frequency representation as input after wavelet transformation.

%\begin{wraptable}{r}{0.58\linewidth}
%\vspace{-25pt}
\begin{table}[t]
\centering
\scriptsize
\setlength{\tabcolsep}{2.5pt}
\renewcommand{\arraystretch}{0.9}

\rowcolors{2}{gray!10}{white}
%\resizebox{\linewidth}{!}{
\begin{tabular}{lccccccc}
\toprule
\textbf{Tokenizer} & \textbf{Rec.}$\downarrow$ & \textbf{LF}$\downarrow$ & \textbf{HF}$\downarrow$ & \textbf{PSNR}$\uparrow$ & \textbf{SSIM}$\uparrow$ & \textbf{LPIPS}$\downarrow$ & \textbf{rFID}$\downarrow$ \\
\midrule
KL-VAE     & 0.1146 & 0.0645 & 0.1313 & 25.07 & 0.74 & 0.2151 & 0.7820 \\
MS-VQ-VAE  & 0.0238 & 0.0705 & 0.0082 & 21.70 & 0.65 & 0.2331 & 0.8427 \\
VA-VAE     & 0.0125 & 0.0263 & 0.0098 & 26.01 & 0.78 & 0.1071 & 0.6509 \\
\textbf{DeBaT} & \textbf{0.0044} & \textbf{0.0114} & \textbf{0.0020} & \textbf{30.03} & \textbf{0.90} & \textbf{0.0940} & \textbf{0.4156} \\
\bottomrule
\end{tabular}
%}
\vspace{1pt}
\caption{Reconstruction comparison of tokenizers using wavelet representations instead of pixel inputs. Metrics include reconstruction loss (Rec.), low-frequency loss (LF), high-frequency loss (HF), PSNR, SSIM, LPIPS, and rFID. Lower is better for Rec., LF, HF, LPIPS, and rFID, while higher is better for PSNR and SSIM.}
\label{tab:reconstruction_ablation}
\vspace{-20pt}
\end{table}
%\end{wraptable}

 Specifically, we retrain these models by coupling the low- and high-frequency components in a shared frequency space, using the same wavelet-based losses employed in our DeBaT framework. In this setup, each tokenizer is optimized jointly on both low- and high-frequency representations obtained via the Discrete Wavelet Transform (DWT) using Haar wavelet filter.

The results, presented in Table~\ref{tab:reconstruction_ablation}, demonstrate that our frequency-aware DeBaT with decoupling frequency representations as input consistently outperforms the coupled-frequency variants across all metrics, including overall reconstruction error (Rec.), low-frequency (LF) and high-frequency (HF) fidelity error, perceptual similarity loss (LPIPS), and reconstruction FID (rFID) with higher PSNR and SSIM. This indicates that explicitly decoupling frequencies during training leads to more accurate and perceptually faithful reconstructions. For a fair comparison, all models are trained with similar combined latent dimensionality as used in the DeBaT configuration. Additional ablations with varying low and high frequency channels are provided in Appendix susection C.4.

\vspace{2mm}

\begin{wraptable}{r}{0.49\linewidth}
\vspace{-15pt}
\centering
\scriptsize
\setlength{\tabcolsep}{2.5pt}
\renewcommand{\arraystretch}{0.9}

\rowcolors{2}{gray!10}{white}
\resizebox{\linewidth}{!}{
\begin{tabular}{@{}lcccc@{}}
\toprule
\textbf{Tokenizer} & \textbf{PSNR}$\uparrow$ & \textbf{gFID}$\downarrow$ & \textbf{Pre.}$\uparrow$ & \textbf{Rec.}$\uparrow$ \\
\midrule
VA-VAE & 33.84 & 6.92 & 0.69 & 0.31 \\
\textbf{DeBaT} & \textbf{34.27} & \textbf{6.86} & \textbf{0.71} & \textbf{0.31} \\
\bottomrule
\end{tabular}
}
\vspace{-6pt}
\caption{\scriptsize High-resolution evaluation on FFHQ ($1024^2$) using LightningDiT. DeBaT achieves better reconstruction and generation quality than VA-VAE demonstrating the scalability to high-resolution generation.}
\label{tab:ffhq_hr}
\vspace{-10pt}
\end{wraptable}

\noindent\textbf{High-resolution evaluation.} We further evaluate DeBaT on FFHQ at $1024^2$ resolution using LightningDiT. As shown in Table~\ref{tab:ffhq_hr}, DeBaT consistently improves both reconstruction quality (PSNR) and generation quality (gFID and precision) over VA-VAE while maintaining the same latent representation ($32\times32\times64$), demonstrating the scalability of the proposed tokenizer to high-resolution generation.

\vspace{2mm}

%\begin{wraptable}{r}{0.6\linewidth}

%\end{wraptable}

\noindent\textbf{Training complexity and efficiency.} Table~\ref{tab:efficiency} compares DeBaT with representative tokenizers in terms of reconstruction quality, architectural complexity, and runtime efficiency. Under the same latent size ($16\times16\times32$) as VA-VAE, DeBaT achieves substantially better reconstruction quality (30.03 dB PSNR vs. 26.26 dB) while maintaining comparable GPU memory consumption (5892 MB vs. 5694 MB) and practical inference speed (50.08 img/s, 19.97 ms/img). Although DeBaT employs more parameters due to its dual encoder--decoder architecture, it remains more efficient than the considerably larger WF-VAE, achieving both higher PSNR and faster inference. These results demonstrate that the proposed frequency-decoupled design provides a favorable trade-off between reconstruction quality and computational efficiency.

\begin{table}[t]
%\vspace{-25pt}
\centering
\scriptsize
\setlength{\tabcolsep}{2.5pt}
\renewcommand{\arraystretch}{0.9}

\rowcolors{2}{gray!10}{white}
%\resizebox{\linewidth}{!}{
\begin{tabular}{lccccc}
\toprule
\textbf{Tokenizer} & \textbf{PSNR}$\uparrow$ & \textbf{\# Params}$\downarrow$ & \textbf{Memory}$\downarrow$ & \textbf{Throughput}$\uparrow$ & \textbf{Latency}$\downarrow$ \\
 &  & (M) & (MB) & (img/s) & (ms/img) \\
\midrule
SD-VAE   & 24.91 & 83.65  & 8040 & 31.78 & 31.47 \\
VA-VAE   & 26.26 & 72.60  & \textbf{5694} & \textbf{57.83} & \textbf{17.29} \\
LiteVAE  & 28.59 & \textbf{59.96} & 6922 & 45.96 & 21.76 \\
WF-VAE   & 29.00 & 317.08 & 9571 & 27.48 & 36.39 \\
\textbf{DeBaT} & \textbf{30.03} & 137.91 & 5892 & 50.08 & 19.97 \\
\bottomrule
\end{tabular}
%}
\vspace{1pt}
\caption{Training complexity and efficiency comparison. Throughput, latency, and GPU memory are measured at batch size 32 and $256\times256$ resolution. Higher PSNR/throughput and lower parameters, memory, and latency are preferred.}
\label{tab:efficiency}
\vspace{-10pt}
\end{table}

%\vspace{-0.6cm}
\section{Conclusion}
In this work, we investigate frequency awareness in the latent embeddings of tokenizer models used within latent generative pipelines. We find that jointly optimizing low- and high-frequency components leads to a frequency bias favoring low-frequencies thereby degrading the reconstruction of fine-grained, high-frequency details and overall perceptual quality. To address this limitation, we introduce DeBaT, a simple yet effective VAE framework that explicitly decouples low and high-frequency bands via wavelet decomposition, processes them independently, and fuses them into a unified latent representation. DeBaT achieves state-of-the-art performance across frequency-aware reconstruction metrics, demonstrating improved fidelity and perceptual quality. When used within a diffusion-based generation framework, our approach preserves the spectral richness of the data, enabling more accurate, high fidelity generation of images with coarse structures as well as fine details.

\noindent\textbf{Discussion \& Future Direction. }
%Our approach DeBaT learns explicit low and high frequency latent embeddings. 
After training, DeBaT's latent code is composed of separate codes for high and low frequencies. A potential future direction is to explore the combination of the learned latent codes with other, complementary latent encoders, where we would hope to improve fine details in generated data without full training. 
Furthermore, it would be interesting to explore DeBaT's use in conjuction with different tokenizer architectures. In this work, we focused on VAVAE as a baseline model because it recently provided state-of-the-art performance while being computationally efficient. 
Last, it would be interesting to extend DeBaT to more wavelet decomposition stages. Our current model only leverages a single wavelet decomposition stage which provides only limited efficiency gains. Multi-Stage decompositions similar to the setting recently proposed in WF-VAE~\cite{li2025wfvae} could be combined with DeBaT, to facilitate fully decoupled multi-stage encoding, potentially with further benefits in efficiency as well as in detail preservation. 
%As , our high frequency latent can be utilized for 

\noindent \textbf{Acknowledgments. } The authors acknowledge support by the state of Baden-Württemberg through bwHPC and the German Research Foundation (DFG) through grant INST 35/1597-1 FUGG.

%\clearpage

%\bibliographystyle{splncs04}
%\bibliography{main}

%\clearpage

\appendix
\begin{appendix}

\title{Decoupling High and Low Frequencies for Faithful Image Generation with Fine Details\\
\Large Appendix}

\titlerunning{DeBaT}

\author{
Tejaswini Medi\inst{1}\orcidlink{0009-0004-6345-2693}
\and
Hsien-Yi Wang\inst{1}
\and
Arianna Rampini\inst{2}
\and
Margret Keuper\inst{1,3}\orcidlink{0000-0002-8437-7993}
}

\authorrunning{T.~Medi et al.}

\institute{University of Mannheim, Germany \and
Autodesk AI Lab \and
MPI for Informatics, Saarland Informatics Campus, Saarbrücken,  Germany \\
\email{\{tejaswini.medi,keuper\}@uni-mannheim.de}}

\maketitle

\noindent In this appendix, we provide additional details supporting the main paper. The following sections elaborate on architectural components of our approach DeBaT, implementation specifics, additional evaluations, qualitative visualizations  and auxiliary analysis that complement our primary findings.

The overview of appendix is provided below:

\begin{itemize}
    \item \textbf{Section~\ref{sec:debat-architecture}}:\;
    Architectural details of the proposed DeBaT.

    \item \textbf{Section~\ref{sec:implementation}}:\;
    Implementation details of DeBaT.

    \item \textbf{Section~\ref{sec:additional_quantitative}}:\;
     Additional quantitative evaluations comparing DeBaT.

    \item \textbf{Section~\ref{sec:qualitative}}:\; Qualitative comparisons with DeBaT, including error maps, residual power spectra, and low/high-frequency reconstructions demonstrating the efficacy of DeBaT. 

   % \item \textbf{Section~\ref{sec:qualitative}}:\;
    %Frequency-aware reconstruction analysis, including error maps, residual power spectra, and low/high-frequency reconstructions demonstrating the efficacy of DeBaT.
\end{itemize}

\section{Architectural Details of DeBaT}
\label{sec:debat-architecture}

The proposed \textbf{Decoupled Band Tokenizer (DeBaT)} separates the learning of low- and high-frequency components in latent tokenizers. Specifically, we first apply a single-level Haar wavelet decomposition to the input data, which decomposes it into low-frequency (LF) and high-frequency (HF) components. Each frequency band is then processed by an independent encoder--decoder pair following a U-Net-based architecture inspired by latent visual tokenizers~\cite{yao2025vavae, esser2021taming}. The low-frequency band focuses on capturing global structural information, while the high-frequency band models fine textures and rapidly varying details. By decoupling the learning process and assigning a dedicated encoder--decoder pair to each frequency band, DeBaT learns disentangled latent representations for low- and high-frequency components. This design enables the model to better exploit the spectral characteristics of each band and improves the representation of both global structure and fine details. After encoding, the latent vectors from the LF and HF are concatenated to form a unified latent representation. This fused latent code is subsequently used by the downstream generative model based on LightningDiT~\cite{yao2025vavae}, enabling frequency-aware image generation.

In this section, we provide pseudocode for the main modules of DeBaT. As illustrated in Figure~\ref{fig:debat-wavelet-modules}, DeBaT employs a wavelet-based splitting mechanism that decomposes the input into low-frequency (LF) and high-frequency (HF) bands, which are later used for reconstruction. The corresponding latent extraction and fusion pipeline is shown in Figure~\ref{fig:debat-latent-fusion}. Finally, Figure~\ref{fig:debat-arch} presents the complete DeBaT pseudo-architecture. The encoder and decoder blocks follow the VAVAE-style encoder–decoder design and are implemented as standard U-Net-based VAEs~\cite{yao2025vavae}. The channel dimensions for the LF and HF branches are configured according to the single wavelet decomposition, which is 16.

\begin{figure*}[t]
\centering
\begin{lstlisting}[style=pythonstyle, basicstyle=\ttfamily\tiny]
class WaveletSplit(nn.Module):
    def __init__(self, mode="haar"):
        super().__init__()
        # 1-level Haar wavelet transform
        self.dwt = DWTForward(J=1, wave=mode, mode="zero")

    def forward(self, x):
        # x: input image (B, 3, H, W)
        x_low, x_high_list = self.dwt(x)   # low and high frequency parts
        x_high = x_high_list[0]            # LH, HL, HH subbands

        # Normalize low / high frequencies
        x_low  = x_low  / 2.0
        x_high = x_high / 2.0

        # Pack three high-frequency subbands into a 9-channel tensor
        B = x_high.shape[0]
        x_high = x_high.reshape(B, 9, x_high.shape[-2], x_high.shape[-1])

        return x_low, x_high


class WaveletMerge(nn.Module):
    def __init__(self, mode="haar"):
        super().__init__()
        # Inverse Haar wavelet transform
        self.idwt = DWTInverse(wave=mode, mode="zero")

    def forward(self, x_low, x_high):
        # x_low:  low-frequency tensor   (B, 3, H/2, W/2)
        # x_high: high-frequency tensor  (B, 9, H/2, W/2)

        # Unpack 9-channel tensor back into three subbands
        B, _, H, W = x_high.shape
        x_high = x_high.reshape(B, 3, 3, H, W)
        x_high = [x_high[:, i] for i in range(3)]  # list: [LH, HL, HH]

        # Undo normalization
        x_low  = x_low  * 2.0
        x_high = [h * 2.0 for h in x_high]

        # Reconstruct full-resolution image
        x_recon = self.idwt((x_low, x_high))
        return x_recon
\end{lstlisting}

\caption{Wavelet decomposition and reconstruction modules used in DeBaT.}
\label{fig:debat-wavelet-modules}
\end{figure*}

\begin{figure*}[t]
\centering
\begin{lstlisting}[style=pythonstyle, basicstyle=\ttfamily\tiny]
# DeBaT: Low- and High-Frequency Latent Extraction
class DeBaT(nn.Module):
    def encode(self, x):
        z_L = self.encoder_L(x)     # low-frequency (global structure)
        z_H = self.encoder_H(x)     # high-frequency (fine details)
        return z_L, z_H

# Fusion Module: z_tilde = F(z_L, z_H)
class LatentFusion(nn.Module):
    def forward(self, z_L, z_H):
        return torch.cat([z_L, z_H], dim=1)

# Integration with LightningDiT for frequency-aware generation
class LightningDiT(nn.Module):
    def __init__(self, DeBaT, dit):
        super().__init__()
        self.DeBaT, self.fusion, self.dit = DeBaT, LatentFusion(), dit

    def forward(self, x, t, cond=None):
        z_L, z_H  = self.DeBaT.encode(x)  # 1) Encode
        z_tilde   = self.fusion(z_L, z_H)  # 2) Fuse
        return self.dit(z_tilde, t, cond)  # 3) Diffusion generation
\end{lstlisting}

\caption{Latent extraction and fusion in DeBaT for frequency-aware generation. The DeBaT encoders independently map low-frequency (\(z_L\)) and high-frequency (\(z_H\)) components into disentangled latent representations. 
The \texttt{LatentFusion} module concatenates the two latent vectors to obtain a unified latent code \(\tilde{z}\), which is then passed to a diffusion transformer (LightningDiT) for frequency-aware generative modeling.}
\label{fig:debat-latent-fusion}
\end{figure*}

\begin{figure*}[t]
\centering
\setlength{\fboxsep}{1pt}

\begin{minipage}[t]{0.50\textwidth}
\begin{lstlisting}[
style=pythonstyle,
basicstyle=\ttfamily\tiny,
lineskip=-1pt,
caption={},
label={lst:wfvae-enc}
]
class DeBaT(nn.Module):
    def __init__(self, encoder_config_L, encoder_config_H,
                 decoder_config_L, decoder_config_H,
                 embed_dim_L, embed_dim_H):
        super().__init__()

        # Wavelet decomposition
        self.wavelet_split = WaveletSplit(mode="haar", levels=1)
        self.wavelet_merge = WaveletMerge(mode="haar")

        # LF encoder/decoder
        self.encoder_L = Encoder(**encoder_config_L)
        self.decoder_L = Decoder(**decoder_config_L)

        # HF encoder/decoder
        self.encoder_H = Encoder(**encoder_config_H)
        self.decoder_H = Decoder(**decoder_config_H)

        # Gaussian posterior parameters
        self.quant_conv_L = nn.Conv2d(
            2 * encoder_config_L["z_channels"],
            2 * embed_dim_L, 1)

        self.quant_conv_H = nn.Conv2d(
            2 * encoder_config_H["z_channels"],
            2 * embed_dim_H, 1)

        # Map latent vectors to decoder channels
        self.post_quant_conv_L = nn.Conv2d(
            embed_dim_L, decoder_config_L["z_channels"], 1)

        self.post_quant_conv_H = nn.Conv2d(
            embed_dim_H, decoder_config_H["z_channels"], 1)

    # LF encoder
    def encode_low(self, x_low):
        h_L = self.encoder_L(x_low)
        moments = self.quant_conv_L(h_L)
        return DiagonalGaussianDistribution(moments)
\end{lstlisting}
\end{minipage}\hfill
\begin{minipage}[t]{0.48\textwidth}
\begin{lstlisting}[
style=pythonstyle,
basicstyle=\ttfamily\tiny,
lineskip=-1pt,
caption={},
label={lst:wfvae-dec}
]

    def encode_high(self, x_high):
        h_H = self.encoder_H(x_high)
        moments = self.quant_conv_H(h_H)
        return DiagonalGaussianDistribution(moments)

    # HF decoders
    def decode_low(self, z_L):
        z_L = self.post_quant_conv_L(z_L)
        return self.decoder_L(z_L)

    def decode_high(self, z_H):
        z_H = self.post_quant_conv_H(z_H)
        return self.decoder_H(z_H)

    # forward
    def forward(self, x):
        x_low, x_high = self.wavelet_split(x)

        q_L = self.encode_low(x_low)
        q_H = self.encode_high(x_high)

        z_L = q_L.sample()
        z_H = q_H.sample()

        x_low_rec  = self.decode_low(z_L)
        x_high_rec = self.decode_high(z_H)

        x_rec = self.wavelet_merge(x_low_rec, x_high_rec)

        return {
            "x_rec": x_rec,
            "x_low_rec": x_low_rec,
            "x_high_rec": x_high_rec,
            "posterior_L": q_L,
            "posterior_H": q_H
        }
\end{lstlisting}
\end{minipage}

\vspace{-0.6em}
\caption{DeBaT pseudo architecture with separate low- and high-frequency encoder–decoder branches.}
\label{fig:debat-arch}
\end{figure*}

\section{Implementation Details}\label{sec:implementation}
We follow a similar hyperparameter configuration to prior works~\cite{rombach2022high, yao2025vavae} for implementing our latent reconstruction and generative modeling pipeline. The visual tokenizer training setup is based on the Latent Diffusion Model (LDM) framework~\cite{rombach2022high, li2024autoregressive}. Specifically, we adopt a VQGAN-style encoder--decoder architecture but remove the vector quantization stage, allowing the model to learn a continuous latent representation regularized with a KL divergence term. The KL loss is applied directly to the latent distribution following~\cite{yao2025vavae}. To support distributed training across multiple nodes, we scale the learning rate and global batch size to $1 \times 10^{-4}$ and 256, respectively, following the MAR training setup~\cite{li2024autoregressive}. For the proposed DeBaT tokenizer, we train two separate $f16$ tokenizers corresponding to the low-frequency and high-frequency bands obtained through wavelet decomposition, where "$f$" represents the latent spatial compression rate. The low-frequency tokenizer is trained using the Visual Foundation (VF) alignment loss introduced in VAVAE~\cite{yao2025vavae}, which leverages features from DINOv2~\cite{oquab2023dinov2}. The VF loss hyperparameters are set to $m_1 = 0.5$, $m_2 = 0.25$, and $w_{\text{hyper}} = 0.1$, following the configuration used in VAVAE~\cite{yao2025vavae}. For generative modeling, we employ LightningDiT~\cite{yao2025vavae}, incorporating the architectural improvements and training strategies described in their work. The latent representations extracted by DeBaT on ImageNet at a resolution of $256 \times 256$ are used to train LightningDiT for 64 epochs. The patch size is set to 1, ensuring that the effective downsampling factor of the overall system remains 16, consistent with the design choice~\cite{chen2024deep, yao2025vavae}. All experiments are conducted using 4 NVIDIA H100 GPUs with 93.6GB VRAM each.

\vspace{0.5em}
\noindent
\textbf{Evaluation Metrics.}
We report several standard metrics commonly used to evaluate the quality of generated images from autoregressive and latent diffusion models. 

\begin{itemize}
    \item \textbf{gFID} (generation FID): Frechet Inception Distance computed between generated images and real images, using the Inception-v3 network. Lower values indicate better alignment in feature distributions and overall visual fidelity.
    \item \textbf{sFID} (spatial FID): A spatial-aware variant of FID. sFID is equivalent to FID but uses intermediate spatial features in the inception network rather than the spatially-pooled features used in standard FID.
    \item \textbf{IS} (Inception Score): Measures both the confidence and diversity of generated images based on class predictions from the Inception network. Higher IS indicates better generative diversity and visual sharpness.
    \item \textbf{Pre. (Precision) and Rec. (Recall)}: These metrics assess the coverage and fidelity of the generated data manifold with respect to real images. Higher precision implies fewer unrealistic samples, while higher recall indicates better diversity and coverage.
\end{itemize}

All metrics are computed using  50K generated samples. We report results both \textbf{with} and \textbf{without Classifier-Free Guidance (CFG)} when available, which is a commonly used sampling technique in diffusion models to improve sample quality. Missing entries (marked as \texttt{-}) reflect the unavailability of results in the respective original papers.

\section{Additional Quantitative Evaluation}
\label{sec:additional_quantitative}
In this section, we provide additional quantitative evaluations to further demonstrate the effectiveness of DeBaT compared to baseline latent tokenizers. We evaluate reconstruction quality using both pixel-level and perceptual metrics. All evaluations are performed on the ImageNet validation set, which consists of 50,000 images. We first analyze the reconstruction fidelity of different visual tokenizers using standard image reconstruction metrics such as PSNR and SSIM. We then evaluate the pixel-level and frequency-aware reconstruction performance of the tokenizers using perceptual similarity metrics and frequency-specific losses for models trained on large-scale datasets. These experiments highlight the advantages of decoupled frequency modeling in DeBaT.

\subsection{Reconstruction Quality: PSNR and SSIM}

We evaluate the reconstruction quality of different visual tokenizers using the standard image reconstruction metrics, Peak Signal-to-Noise Ratio (PSNR) and Structural Similarity Index (SSIM). These metrics measure pixel-level reconstruction fidelity and structural consistency between the original and reconstructed images.

%Table~\ref{tab:psnr_ssim_comparison} reports the PSNR and SSIM scores for various latent tokenizer models. The results demonstrate that DeBaT achieves high PSNR and SSIM compared to other latent tokenizers leading to improved reconstruction fidelity, highlighting the benefits of decoupling low-frequency structural information from high-frequency details during encoding and decoding.

Table~\ref{tab:psnr_ssim_comparison} reports PSNR and SSIM scores for various latent tokenizer models. The results show that DeBaT achieves higher PSNR and SSIM compared to other latent tokenizers, leading to improved reconstruction fidelity. These results highlight the benefit of decoupling low-frequency structural information from high-frequency details during encoding and decoding. We additionally include WF-VAE~\cite{li2025wfvae}, a wavelet-based tokenizer originally designed for video reconstruction, in the comparison. To ensure fair evaluation in the image setting, we adapt WF-VAE to operate on single-frame inputs (T=1) using the image training pipeline with comparable optimization settings. Despite this adaptation, DeBaT consistently achieves better reconstruction performance.

\begin{table*}[t]
\centering
\scriptsize
\setlength{\tabcolsep}{6pt}
\renewcommand{\arraystretch}{0.9}
\rowcolors{2}{gray!10}{white}

\begin{tabular}{llcc}
\toprule
\textbf{Visual Tokenizer} & \textbf{Tokenizer Latent} & \textbf{PSNR}$(\uparrow)$ & \textbf{SSIM}$(\uparrow)$ \\
\midrule
DC-AE~\cite{chen2024deep} & 8x8x32 & 23.75 & 0.69 \\
DC-AE~\cite{chen2024deep} & 4x4x128 & 23.47 & 0.69 \\
DC-AE~\cite{chen2024deep} & 4x4x512 & 23.07 & 0.66 \\
SD-VAE~\cite{rombach2022high} & 16x16x16 & 24.91 & 0.74 \\
KL-VAE~\cite{li2024autoregressive} & 16x16x16 & 25.76 & 0.75 \\
MS-VQ-VAE~\cite{tian2024visual} & 16x16x32 (4096) & 22.58 & 0.67 \\
RQ-VAE~\cite{lee2022autoregressive} & 8x8x16 (16384) & 20.11 & 0.54 \\
TiTok-VQ-VAE~\cite{yu2024image} & 128x64 (8192) & 24.02 & 0.67 \\
TiTok-VQ-VAE~\cite{yu2024image} & 32x64 (8192) & 20.72 & 0.53 \\
TiTok-VAE~\cite{yu2024image} & 256x16 & 22.03 & 0.60 \\
TiTok-VAE~\cite{yu2024image} & 128x16 & 20.46 & 0.53 \\
TiTok-VAE~\cite{yu2024image} & 64x16 & 19.07 & 0.47 \\
VQ-VAE~\cite{esser2021taming} & 16x16x256 (1024) & 19.44 & 0.52\\
VQ-VAE~\cite{esser2021taming} & 16x16x256 (16384) & 19.96 & 0.54 \\
WF-VAE~\cite{li2025wfvae} & 32x32x16 &  29.00 & 0.88 \\
LiteVAE~\cite{sadat2024litevae} & 32x32x12 & 28.59 & 0.86 \\
VA-VAE~\cite{yao2025vavae} & 16x16x32 & 26.26 & 0.80 \\
\textbf{{DeBaT (Ours)}} & {\textbf{16x16x32}} & {\textbf{30.03}} & {\textbf{0.90}} \\
\bottomrule
\end{tabular}
\vspace{1pt}
\caption{Additional quantitative comparison of the visual tokenizers from Table 1 in main paper based on reconstruction fidelity measured using \textbf{PSNR} and \textbf{SSIM} (higher is better). These tokenizers are trained on the ImageNet-1k training set and evaluated on the validation set. Tokenizer latent denotes the latent dimensionality (height $\times$ width $\times$ channels). For quantized tokenizers, the codebook size is shown in parentheses. For TiTok variants with a 1D latent space, the latent dimension is represented as (number of tokens $\times$ channels). Higher PSNR and SSIM indicate better reconstruction quality. Wavelet-based tokenizers such as LiteVAE and our DeBaT substantially outperform other methods in these metrics. DeBat performs best. }
\label{tab:psnr_ssim_comparison}
\end{table*}

\subsection{Frequency-aware and Perceptual Evaluation}

To further analyze the reconstruction quality of our approach in comparison with visual tokenizers trained on large-scale datasets, we evaluate several such tokenizers alongside our proposed framework, \textbf{DeBaT}. The comparison focuses on both perceptual and frequency-aware reconstruction metrics. Specifically, we report commonly used image reconstruction metrics such as PSNR, SSIM, and reconstruction loss, together with frequency-sensitive measures including low-frequency loss and high-frequency loss. In addition, we evaluate perceptual similarity using the LPIPS metric.

%To further analyze reconstruction quality of our approach compared to visual tokenizers which are trained on large-scale datasets, We compare various visual tokenizers trained on large-scale datasets  with respect to our framework DeBaT. we evaluate the tokenizers using perceptual and frequency-aware metrics. Specifically, we report standard metrics like PSNR, SSIM, reconstruction loss along with low-frequency loss, high-frequency loss, and LPIPS perceptual similarity across different tokenizers.

%Table~\ref{tab:semantic-comparison} presents a comparison of these metrics across multiple latent tokenizers and DeBaT. The results show that DeBaT consistently achieves better frequency-aware reconstruction while maintaining strong perceptual performance, demonstrating the effectiveness of decoupled frequency learning in latent tokenization.

\begin{table*}[t]
\centering
\scriptsize
\setlength{\tabcolsep}{4pt}
\renewcommand{\arraystretch}{0.9}
\rowcolors{2}{gray!10}{white}
\begin{tabular}{lp{2.0cm}cccccc}
\toprule
\textbf{Tokenizer} & \textbf{Tokenizer Config} & \textbf{PSNR} & \textbf{SSIM} & \textbf{Recon.} & \textbf{LF. Loss} & \textbf{HF. Loss} & \textbf{LPIPS} \\
\midrule
DC-AE$^*$           & 8x8x32      & 23.32 & 0.68 & 0.0213 & 0.0521 & 0.0110 & 0.1708 \\
DC-AE$^*$           & 4x4x128      & 23.47 & 0.69 & 0.0206 & 0.0499 & 0.0109 & 0.1689 \\
DC-AE$^*$           & 4x4x512     & 23.43 & 0.68 & 0.0208 & 0.0510 & 0.0107 & 0.1756 \\
KL-VAE$^\dagger$    & 64x64x3          & 28.63 & 0.86 & 0.0064 & 0.0097 & 0.0053 & 0.0722 \\
KL-VAE$^\dagger$    & 32x32x4          & 24.67 & 0.73 & 0.0156 & 0.0348 & 0.0091 & 0.1459 \\
KL-VAE$^\dagger$    & 16x16x16        & 24.46 & 0.72 & 0.0163 & 0.0365 & 0.0096 & 0.1529 \\
KL-VAE$^\dagger$    & 8x8x64        & 22.50 & 0.64 & 0.0251 & 0.0640 & 0.0121 & 0.2089 \\
KL-VAE$^\dagger$    & 4x4x128       & 23.10 & 0.66 & 0.0241 & 0.0652 & 0.0105 & 0.1708 \\
VQ-VAE$^\dagger$    & 64x64x3(8192)     & 28.57 & 0.86 & 0.0064 & 0.0104 & 0.0051 & 0.0693 \\
VQ-VAE$^\dagger$    & 32x32x4(16384)    & 23.43 & 0.69 & 0.0203 & 0.0475 & 0.0113 & 0.1701 \\
VQ-VAE$^\dagger$    & 32x32x4(256)      & 22.70 & 0.66 & 0.0239 & 0.0569 & 0.0128 & 0.1848 \\
VQ-VAE$^\dagger$    & 16x16x8(16384)   & 20.94 & 0.57 & 0.0352 & 0.1006 & 0.0134 & 0.2637 \\
KL-VAE$^\ddagger$   & 32x32x4(ema)    & 25.22 & 0.75 & 0.0145 & 0.0313 & 0.0089 & 0.1345 \\
%\rowcolor{green!8}
\textbf{{DebaT}} & \textbf{{16x16x32}} & \textbf{{30.03}} & \textbf{{0.90}} & \textbf{{0.0044}} & \textbf{{0.0114}} & \textbf{{0.0020}} & \textbf{{0.0940}} \\
\bottomrule
\end{tabular}
\vspace{1pt}
\caption{Comparison of DeBaT (trained only on ImageNet-1k) to latent tokenizers from related work trained on large-scale datasets. All models are evaluated on the ImageNet validation set across pixel-level, frequency-aware, and perceptual metrics. The tokenizer configuration column reports the latent spatial resolution and channel count, shown as $h{\times}w{\times}c$; for VQ-based models, the vocabulary size $v$ is shown in parentheses. Higher values are better for \textbf{PSNR} and \textbf{SSIM}, while lower values indicate better performance for reconstruction loss, low-frequency loss, high-frequency loss, and LPIPS. $^*$ Models trained on ImageNet, SAM, FFHQ, and Mapillary Vistas. $^\dagger$ Models trained on OpenImages. $^\ddagger$ Models fine-tuned on OpenImages and LAION.}
%\caption{Comparison of DeBaT (trained only on ImageNet-1k) to Latent tokenizers from related work that are trained on large-scale datasets. All are evaluated on the ImageNet validation set across pixel-level, frequency-aware, and perceptual metrics. Higher values are better for \textbf{PSNR} and \textbf{SSIM}, while lower values indicate better performance for reconstruction loss, low-frequency loss, high-frequency loss, and LPIPS. \textbf{f} denotes the latent spatial downsampling factor (e.g., $f{=}16$ implies $16\times$ downsampling in width and height), and \textbf{c} denotes the number of latent channels. For VQ-based models, \textbf{v} denotes the vocabulary size. $^*$ Models trained on ImageNet, SAM, FFHQ, and Mapillary Vistas. $^\dagger$ Models trained on OpenImages. $^\ddagger$ Models fine-tuned on OpenImages and LAION.}
\label{tab:semantic-comparison}
\end{table*}

\begin{figure*}[th!]
\centering\vspace{-2mm}
\begin{subfigure}[b]{0.135\textwidth}
    \centering
    \textbf{\scriptsize Input}\\[2pt]
    \includegraphics[width=\linewidth]{teaser/Fig4/input_00003.png}
    \caption*{\scriptsize Recon. Error:}
\end{subfigure}
\hfill
\begin{subfigure}[b]{0.135\textwidth}
    \centering
    \textbf{\scriptsize WF-VAE}\\[2pt]
    \includegraphics[width=\linewidth]{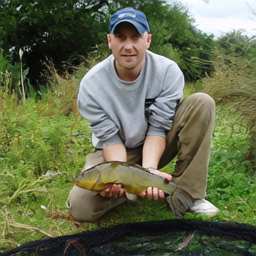}
    \caption*{\scriptsize 0.0025}
\end{subfigure}
\hfill
\begin{subfigure}[b]{0.135\textwidth}
    \centering
    \textbf{\scriptsize LiteVAE}\\[2pt]
    \includegraphics[width=\linewidth]{teaser/Fig4/recon_00003_litevae.png}
    \caption*{\scriptsize 0.0028}
\end{subfigure}
\hfill
\begin{subfigure}[b]{0.135\textwidth}
    \centering
    \textbf{\scriptsize Ours }\\[2pt]
    \includegraphics[width=\linewidth]{teaser/Fig4/recon_00003_ours.png}
    \caption*{\scriptsize 0.0018}
\end{subfigure}
\hfill
\begin{subfigure}[b]{0.135\textwidth}
    \centering
    \textbf{\scriptsize WF-VAE}\\[2pt]
    \includegraphics[width=\linewidth]{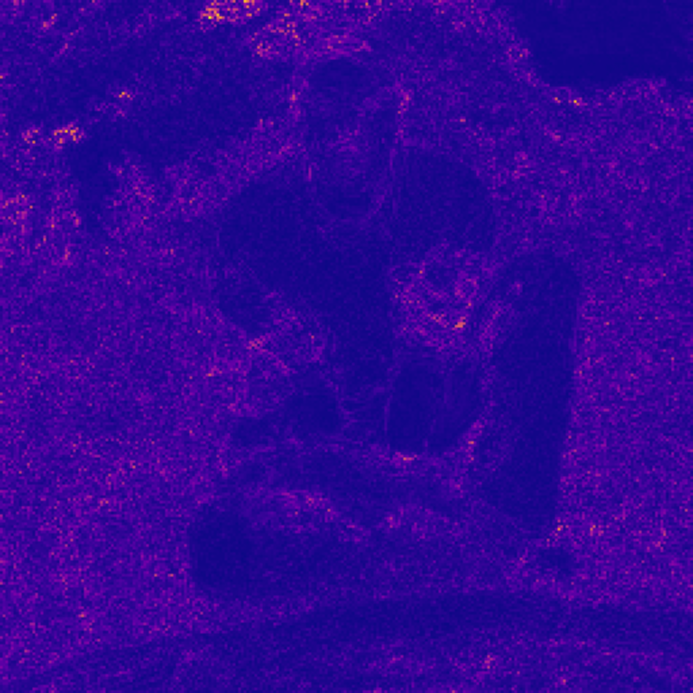}
    \caption*{\scriptsize Error Map}
\end{subfigure}
\hfill
\begin{subfigure}[b]{0.135\textwidth}
    \centering
    \textbf{\scriptsize LiteVAE}\\[2pt]
    \includegraphics[width=\linewidth]{teaser/Fig4/00003_error_litevae.png}
    \caption*{\scriptsize Error Map}
\end{subfigure}
\hfill
\begin{subfigure}[b]{0.135\textwidth}
    \centering
    \textbf{\scriptsize Ours}\\[2pt]
    \includegraphics[width=\linewidth]{teaser/Fig4/00003_error_ours.png}
    \caption*{\scriptsize Error Map}
\end{subfigure}

%%%%%%%%%%%%%%%%%%%%%%%%%%%%%%%%%%%%%%%%%%%%%%%%%%%%%%%%%%%%%

\begin{subfigure}[b]{0.135\textwidth}
    \centering
   % \textbf{Input}\\[2pt]
    \includegraphics[width=\linewidth]{teaser/Fig4/input_30730.png}
    \caption*{\scriptsize Recon. Error:}
\end{subfigure}
\hfill
\begin{subfigure}[b]{0.135\textwidth}
    \centering
  %  \textbf{VA-VAE}\\[2pt]
    \includegraphics[width=\linewidth]{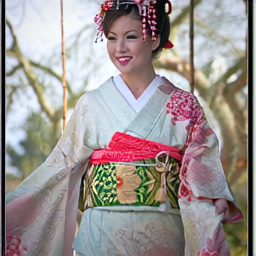}
    \caption*{\scriptsize 0.0024}
\end{subfigure}
\hfill
\begin{subfigure}[b]{0.135\textwidth}
    \centering
   % \textbf{WF-VAE (Ours)}\\[2pt]
    \includegraphics[width=\linewidth]{teaser/Fig4/recon_30730_litevae.png}
    \caption*{\scriptsize 0.0023}
\end{subfigure}
\hfill
\begin{subfigure}[b]{0.135\textwidth}
    \centering
  %  \textbf{VA-VAE Error Map}\\[2pt]
    \includegraphics[width=\linewidth]{teaser/Fig4/recon_30730_ours.png}
    \caption*{\scriptsize 0.0018}
\end{subfigure}
\hfill
\begin{subfigure}[b]{0.135\textwidth}
    \centering
  %  \textbf{VA-VAE Error Map}\\[2pt]
    \includegraphics[width=\linewidth]{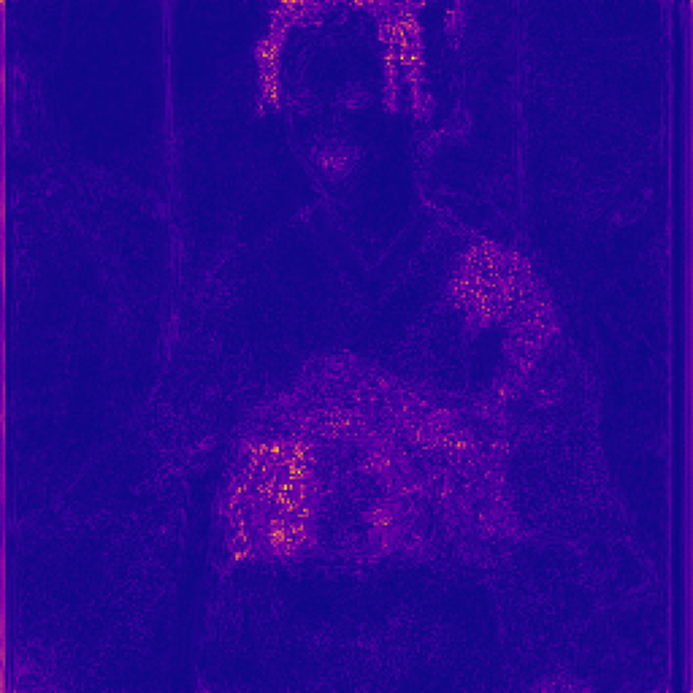}
    \caption*{\scriptsize Error Map}
\end{subfigure}
\hfill
\begin{subfigure}[b]{0.135\textwidth}
    \centering
  %  \textbf{VA-VAE Error Map}\\[2pt]
    \includegraphics[width=\linewidth]{teaser/Fig4/30730_error_litevae.png}
    \caption*{\scriptsize Error Map}
\end{subfigure}
\hfill
\begin{subfigure}[b]{0.135\textwidth}
    \centering
   % \textbf{WF-VAE Error Map}\\[2pt]
    \includegraphics[width=\linewidth]{teaser/Fig4/30730_error_ours.png}
    \caption*{\scriptsize Error Map}
\end{subfigure}

% ==========================================
% ======= Example 3 (00003) ================
% ==========================================
\begin{subfigure}[b]{0.135\textwidth}
    \centering
   % \textbf{Input}\\[2pt]
    \includegraphics[width=\linewidth]{teaser/Fig4/input_31879.png}
    \caption*{\scriptsize Recon. Error:}
\end{subfigure}
\hfill
\begin{subfigure}[b]{0.135\textwidth}
    \centering
  %  \textbf{VA-VAE}\\[2pt]
    \includegraphics[width=\linewidth]{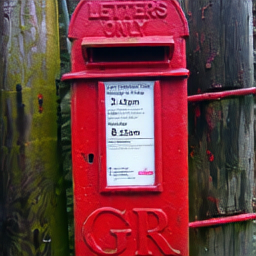}
    \caption*{\scriptsize 0.0016}
\end{subfigure}
\hfill
\begin{subfigure}[b]{0.135\textwidth}
    \centering
   % \textbf{WF-VAE (Ours)}\\[2pt]
    \includegraphics[width=\linewidth]{teaser/Fig4/recon_31879_litevae.png}
    \caption*{\scriptsize 0.0015}
\end{subfigure}
\hfill
\begin{subfigure}[b]{0.135\textwidth}
    \centering
  %  \textbf{VA-VAE Error Map}\\[2pt]
    \includegraphics[width=\linewidth]{teaser/Fig4/recon_31879_ours.png}
    \caption*{\scriptsize 0.0012}
\end{subfigure}
\hfill
\begin{subfigure}[b]{0.135\textwidth}
    \centering
  %  \textbf{VA-VAE Error Map}\\[2pt]
    \includegraphics[width=\linewidth]{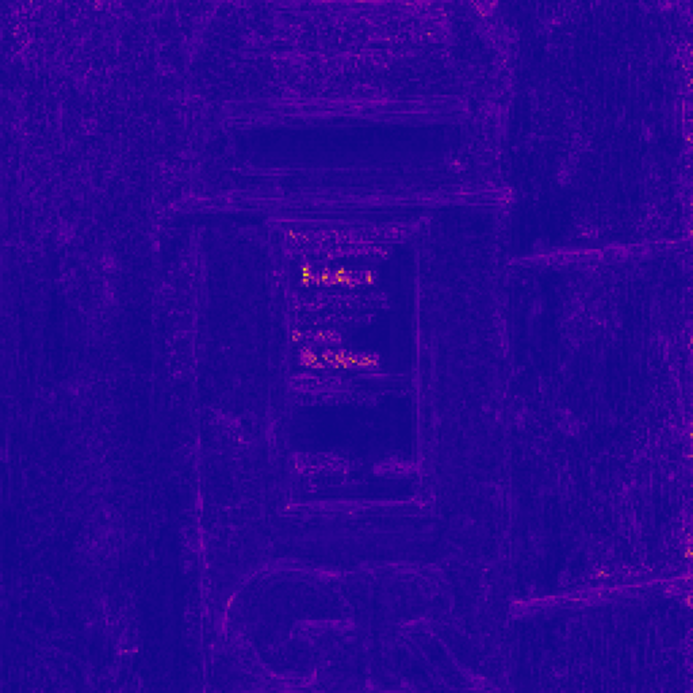}
    \caption*{\scriptsize Error Map}
\end{subfigure}
\hfill
\begin{subfigure}[b]{0.135\textwidth}
    \centering
  %  \textbf{VA-VAE Error Map}\\[2pt]
    \includegraphics[width=\linewidth]{teaser/Fig4/31879_error_litevae.png    }
    \caption*{\scriptsize Error Map}
\end{subfigure}
\hfill
\begin{subfigure}[b]{0.135\textwidth}
    \centering
   % \textbf{WF-VAE Error Map}\\[2pt]
    \includegraphics[width=\linewidth]{teaser/Fig4/31879_error_ours.png}
    \caption*{\scriptsize Error Map}
\end{subfigure}

% ==========================================
% ======= Example 5 (00005) ================
% ==========================================
\begin{subfigure}[b]{0.135\textwidth}
    \centering
   % \textbf{Input}\\[2pt]
    \includegraphics[width=\linewidth]{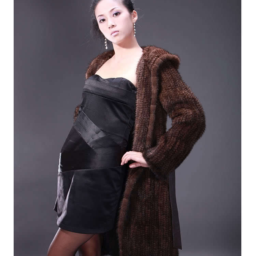}
    \caption*{\scriptsize Recon. Error:}
\end{subfigure}
\hfill
\begin{subfigure}[b]{0.135\textwidth}
    \centering
  %  \textbf{VA-VAE}\\[2pt]
    \includegraphics[width=\linewidth]{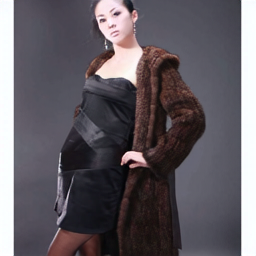}
    \caption*{\scriptsize 0.0002}
\end{subfigure}
\hfill
\begin{subfigure}[b]{0.135\textwidth}
    \centering
   % \textbf{WF-VAE (Ours)}\\[2pt]
    \includegraphics[width=\linewidth]{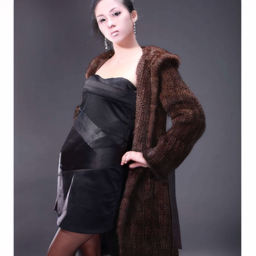}
    \caption*{\scriptsize 0.0002}
\end{subfigure}
\hfill
\begin{subfigure}[b]{0.135\textwidth}
    \centering
  %  \textbf{VA-VAE Error Map}\\[2pt]
    \includegraphics[width=\linewidth]{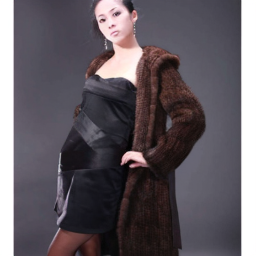}
    \caption*{\scriptsize 0.0001}
\end{subfigure}
\hfill
\begin{subfigure}[b]{0.135\textwidth}
    \centering
  %  \textbf{VA-VAE Error Map}\\[2pt]
    \includegraphics[width=\linewidth]{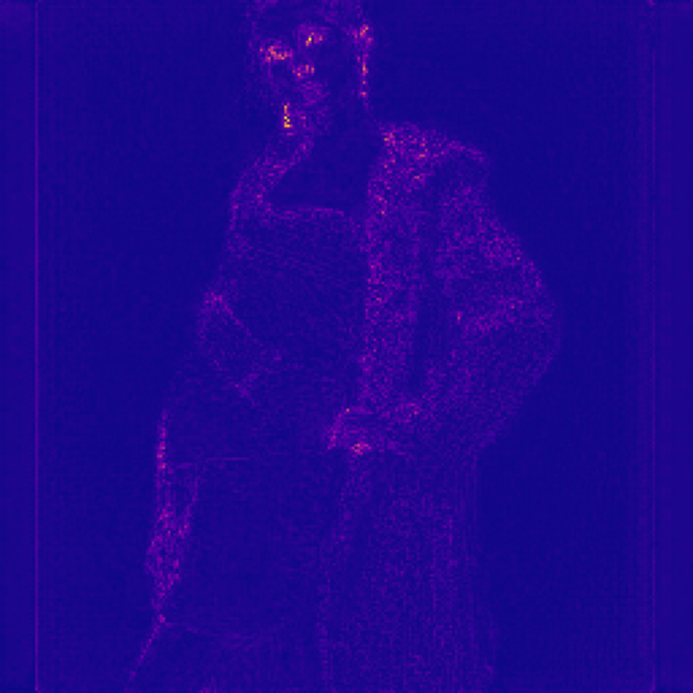}
    \caption*{\scriptsize Error Map}
\end{subfigure}
\hfill
\begin{subfigure}[b]{0.135\textwidth}
    \centering
  %  \textbf{VA-VAE Error Map}\\[2pt]
    \includegraphics[width=\linewidth]{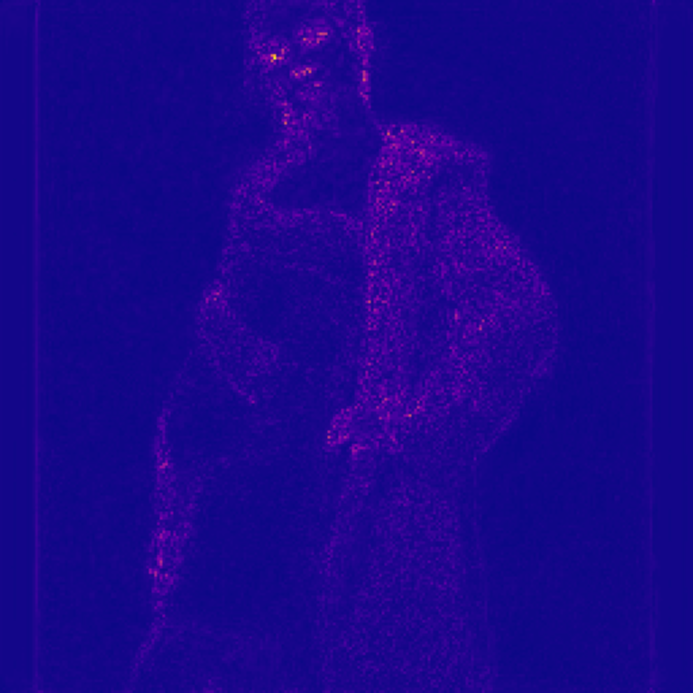  }
    \caption*{\scriptsize Error Map}
\end{subfigure}
\hfill
\begin{subfigure}[b]{0.135\textwidth}
    \centering
   % \textbf{WF-VAE Error Map}\\[2pt]
    \includegraphics[width=\linewidth]{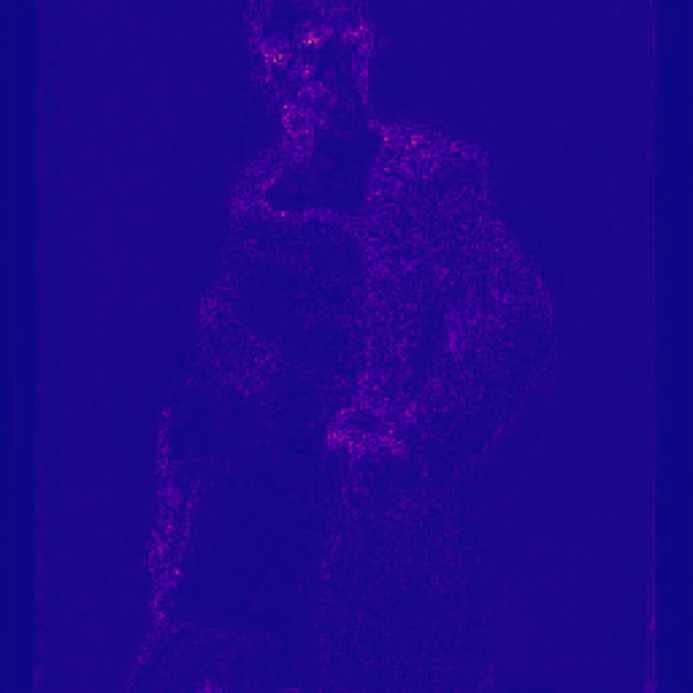}
    \caption*{\scriptsize Error Map}
\end{subfigure}
% ==========================================
% ======= Example 4 (00004) ================
% ==========================================
\begin{subfigure}[b]{0.135\textwidth}
    \centering
   % \textbf{Input}\\[2pt]
    \includegraphics[width=\linewidth]{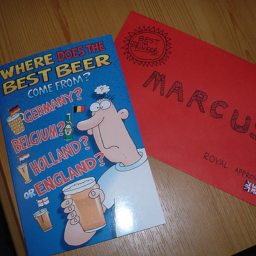}
    \caption*{\scriptsize Recon. Error:}
\end{subfigure}
\hfill
\begin{subfigure}[b]{0.135\textwidth}
    \centering
  %  \textbf{VA-VAE}\\[2pt]
    \includegraphics[width=\linewidth]{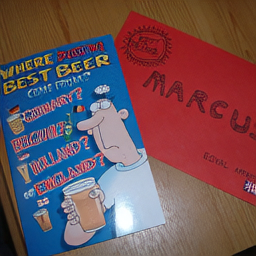}
    \caption*{\scriptsize 0.0029}
\end{subfigure}
\hfill
\begin{subfigure}[b]{0.135\textwidth}
    \centering
   % \textbf{WF-VAE (Ours)}\\[2pt]
    \includegraphics[width=\linewidth]{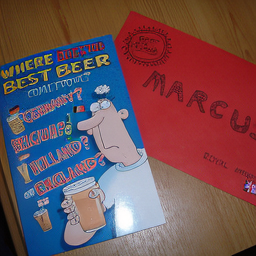}
    \caption*{\scriptsize 0.0030}
\end{subfigure}
\hfill
\begin{subfigure}[b]{0.135\textwidth}
    \centering
  %  \textbf{VA-VAE Error Map}\\[2pt]
    \includegraphics[width=\linewidth]{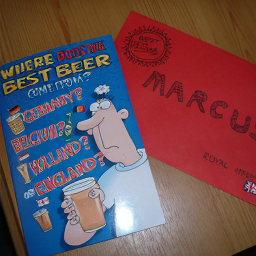}
    \caption*{\scriptsize 0.0022}
\end{subfigure}
\hfill
\begin{subfigure}[b]{0.135\textwidth}
    \centering
  %  \textbf{VA-VAE Error Map}\\[2pt]
    \includegraphics[width=\linewidth]{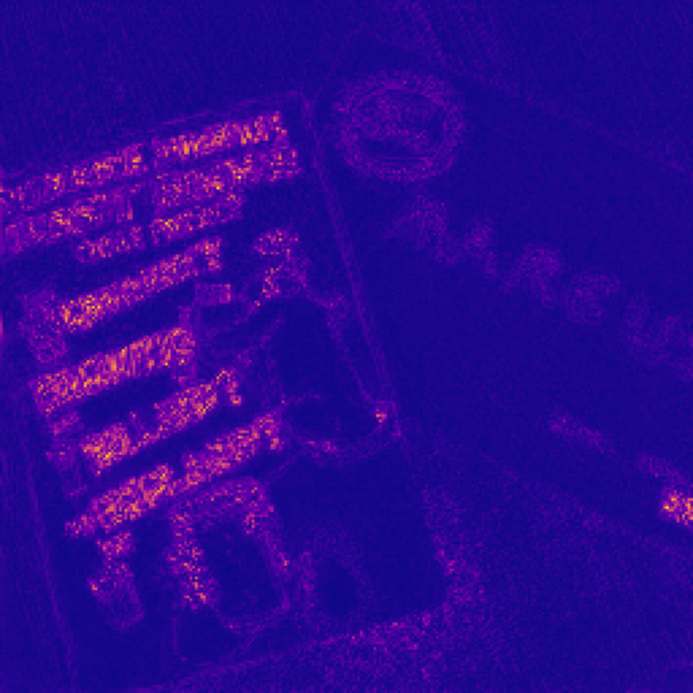}
    \caption*{\scriptsize Error Map}
\end{subfigure}
\hfill
\begin{subfigure}[b]{0.135\textwidth}
    \centering
  %  \textbf{VA-VAE Error Map}\\[2pt]
    \includegraphics[width=\linewidth]{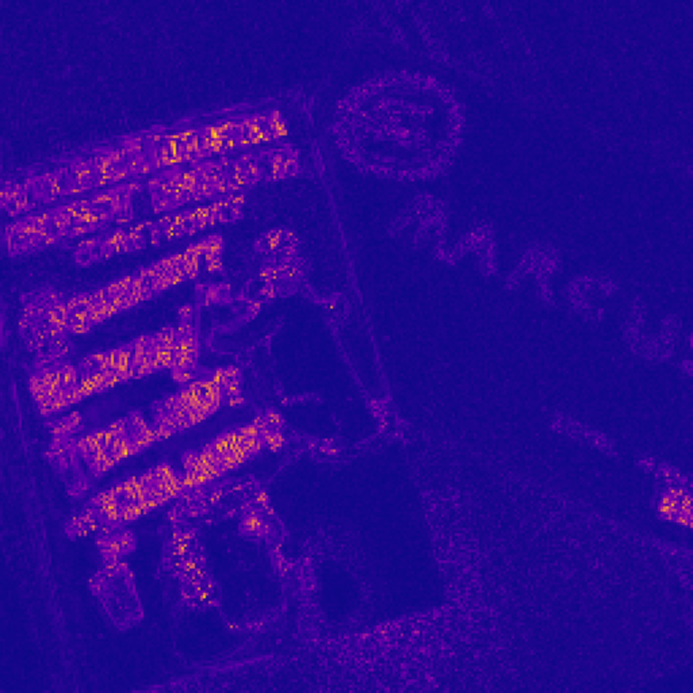  }
    \caption*{\scriptsize Error Map}
\end{subfigure}
\hfill
\begin{subfigure}[b]{0.135\textwidth}
    \centering
   % \textbf{WF-VAE Error Map}\\[2pt]
    \includegraphics[width=\linewidth]{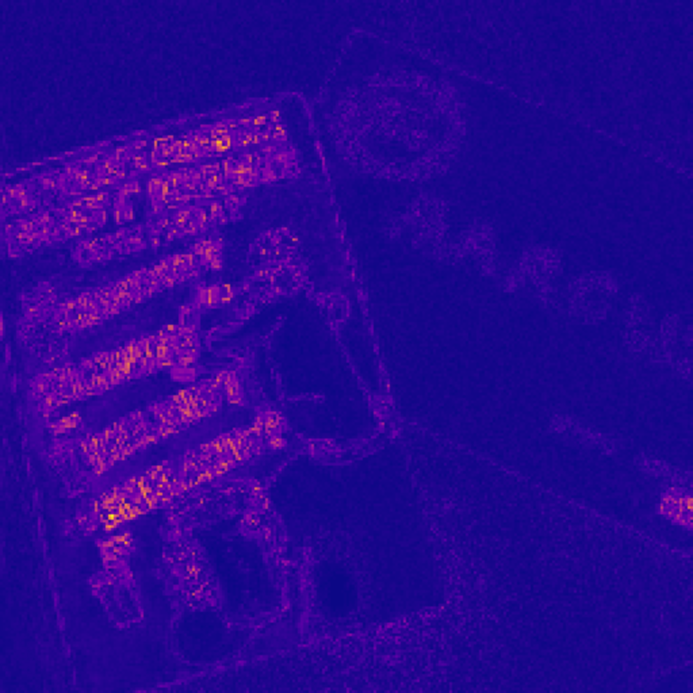}
    \caption*{\scriptsize Error Map}
\end{subfigure}

\vspace{-2mm}

% ====================================================
% ======= Repeat the above block for 00003 ... 00009 =
% ====================================================

% (You will repeat the same structure for examples 00003--00009)
% Just replace file names accordingly!

\caption{Comparison of DeBaT reconstructions with other wavelet based approaches LiteVAE and WF-VAE. DeBaT have consistently lower values of reconstruction error than WF-VAE and LiteVAE, visible as fine details that are correctly reconstructed in particular in structured regions like persons' faces, the texture of the dress or the printed text.}
\label{fig:wave_comparisons}
\end{figure*}

%Table~\ref{tab:semantic-comparison} reports results across both frequency-aware metrics (reconstruction, low-frequency, and high-frequency loss) and LPIPS perceptual similarity of different tokenizers in comparision to DeBaT.
Table~\ref{tab:semantic-comparison} summarizes the results across multiple latent tokenizers and our proposed DeBaT model. The results indicate that DeBaT consistently achieves superior frequency-aware reconstruction while maintaining strong perceptual quality. These findings highlight the effectiveness of decoupled frequency learning for latent tokenization. Our proposed DeBaT achieves the strongest performance across all metrics, demonstrating superior preservation of both coarse structure and fine high-frequency details. It also ranks competitively in perceptual similarity, outperforming prior models such as KL-VAE$^\dagger$ and VQ-VAE$^\dagger$ trained on OpenImages~\cite{krasin2017openimages}, as well as DC-AE$^*$ models trained on broader multi-domain datasets including ImageNet~\cite{deng2009imagenet}, SAM~\cite{kirillov2023segmentanything}, FFHQ~\cite{karras2019style}, and Mapillary Vistas~\cite{neuhold2017mapillary}. We also emphasize that DeBaT is trained only on the ImageNet dataset, whereas other baselines rely here on substantially larger training sets. Given its strong performance under this constraint, we expect that training DeBaT on larger and more diverse datasets would lead to even greater improvements.

The DC-AE, KL-VAE, and VQ-VAE architectures and training setups follow the implementations introduced in respective prior work~\cite{xie2024sana,xie2025sana,rombach2022high}.

\subsection{Longer Generation Training}
\label{sec:longer_generation_training}

Following standard DiT-based ImageNet protocols, prior work commonly trains for 400K iterations with a batch size of 256, corresponding to roughly 80 epochs, and reports continued improvements with extended training~\cite{peebles2023scalable}. We therefore use 400K iterations as a reference point and further evaluate DeBaT under a longer training schedule. Table~\ref{tab:longer_training} compares DeBaT with VA-VAE after 128 epochs on ImageNet at $256{\times}256$ resolution.

\begin{table}[t]
\centering
\scriptsize
\resizebox{\linewidth}{!}{
\begin{tabular}{lcccccccc}
\toprule
Dataset & Res. & Epochs & Latent Dim. & Tokenizer & PSNR$\uparrow$ & gFID-50K$\downarrow$ & Pre.$\uparrow$ & Rec.$\uparrow$ \\
\midrule
ImageNet & $256{\times}256$ & 128 & $16{\times}16{\times}32$ & VA-VAE & 26.26 & 4.95 & 0.76 & 0.62 \\
ImageNet & $256{\times}256$ & 128 & $16{\times}16{\times}32$ & \textbf{DeBaT} & \textbf{30.03} & \textbf{3.09} & \textbf{0.83} & \textbf{0.69} \\
\bottomrule
\end{tabular}
}
\vspace{1pt}
\caption{Longer generation training on ImageNet at $256{\times}256$ resolution. DeBaT maintains its advantage over VA-VAE after 128 epochs, improving both reconstruction fidelity and generation quality.}
\label{tab:longer_training}
\end{table}

At 128 epochs, DeBaT improves generation quality over VA-VAE, reducing gFID-50K from 4.95 to 3.09 while increasing precision and recall from 0.76/0.62 to 0.83/0.69. These results indicate that DeBaT's frequency-decoupled representation remains beneficial under longer generation training.

\subsection{Ablation on Low- and High-Frequency Channel Allocation}
\label{sec:ablation_channels}
%\begin{wraptable}{r}{0.53\textwidth}
%\vspace{-6mm}
\begin{table}[t]
\centering
\scriptsize
\begin{tabular}{l@{\hspace{3mm}}c@{\hspace{3mm}}c@{\hspace{3mm}}c@{\hspace{3mm}}c@{\hspace{3mm}}c@{\hspace{3mm}}c}
\toprule
Configuration & $C_L$ & $C_H$ & rFID$\downarrow$ & LPIPS$\downarrow$ & PSNR$\uparrow$ & SSIM$\uparrow$ \\
\midrule
Standard & 16 & 16 & 0.21 & 0.094 & 30.03 & 0.90 \\
More LF & 32 & 16 & 0.17 & 0.067 & 31.09 & 0.91 \\
More HF & 16 & 32 & 0.21 & 0.087 & 30.84 & 0.90 \\
Both & 32 & 32 & \textbf{0.15} & \textbf{0.057} & \textbf{31.83} & \textbf{0.91} \\
\bottomrule
\end{tabular}
\vspace{1pt}
\caption{Ablation of latent channel allocation between the low-frequency ($C_L$) and high-frequency ($C_H$) branches. Increasing the capacity of the low-frequency branch provides larger gains than increasing the high-frequency branch alone, while enlarging both can boost reconstruction quality further.}
\label{tab:channel_ablation}
%\vspace{-3mm}
\end{table}

The main paper reports the default channel configuration used by DeBaT, where both the low-frequency (LF) and high-frequency (HF) branches employ latent representations of size $16 \times 16 \times 16$. To further investigate the effect of channel allocation, we perform an ablation by varying the number of latent channels assigned to the low-frequency branch ($C_L$) and high-frequency branch ($C_H$). We evaluate four configurations while keeping all other training settings unchanged. we observe that increasing the capacity of the LF branch provides larger gains than increasing the HF branch alone, while enlarging both branches achieves the best overall performance.

Table~\ref{tab:channel_ablation} shows that allocating additional latent channels to the low-frequency branch yields larger improvements than increasing the capacity of the high-frequency branch alone, reducing both rFID and LPIPS while improving PSNR. Increasing the number of channels in both branches further enhances reconstruction quality, achieving the best performance across all evaluation metrics. These results suggest that allocating sufficient capacity to the low-frequency representation is particularly important for preserving global image structure, whereas increasing both branches provides complementary benefits at the cost of increased latent dimensionality.

\begin{figure*}[t]
\centering

% =======================
% ======== ROW 1 ========
% =======================
\begin{subfigure}[t]{0.48\textwidth}
    \centering
    \includegraphics[width=\linewidth]{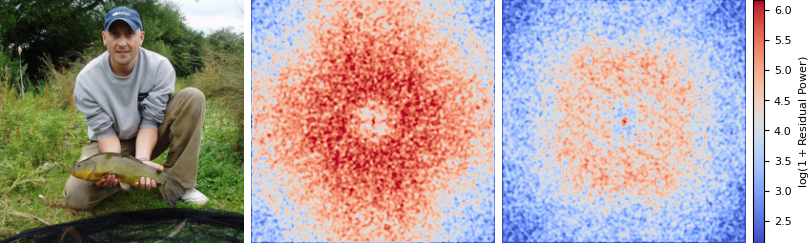}
   % \caption*{\small Input 1}
\end{subfigure}
\hfill
\begin{subfigure}[t]{0.48\textwidth}
    \centering
    \includegraphics[width=\linewidth]{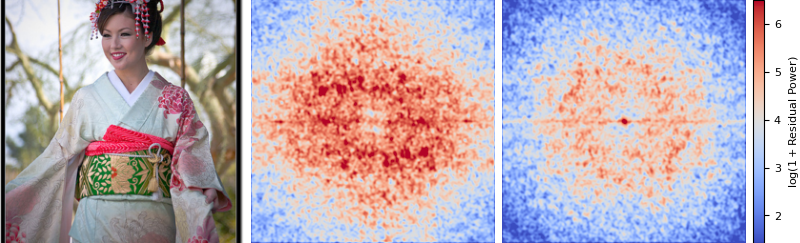}
   % \caption*{\small Input 2}
\end{subfigure}

\vspace{6pt}

% =======================
% ======== ROW 2 ========
% =======================
\begin{subfigure}[t]{0.48\textwidth}
    \centering
    \includegraphics[width=\linewidth]{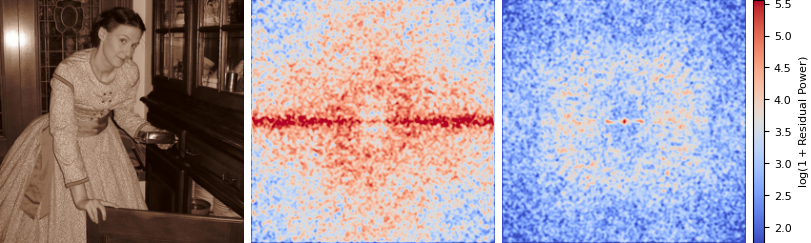}
   % \caption*{\small VAVAE Spectrum (Example 1)}
\end{subfigure}
\hfill
\begin{subfigure}[t]{0.48\textwidth}
    \centering
    \includegraphics[width=\linewidth]{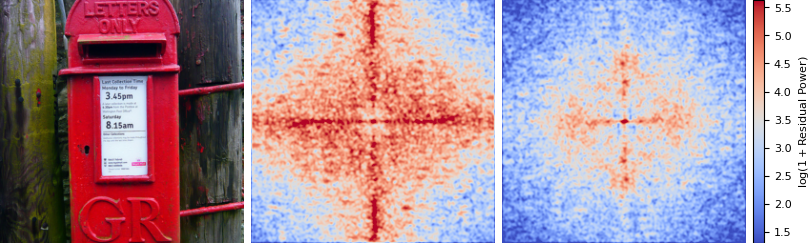}
    %\caption*{\small VAVAE Spectrum (Example 2)}
\end{subfigure}

\vspace{6pt}

% =======================
% ======== ROW 3 ========
% =======================
\begin{subfigure}[t]{0.48\textwidth}
    \centering
    \includegraphics[width=\linewidth]{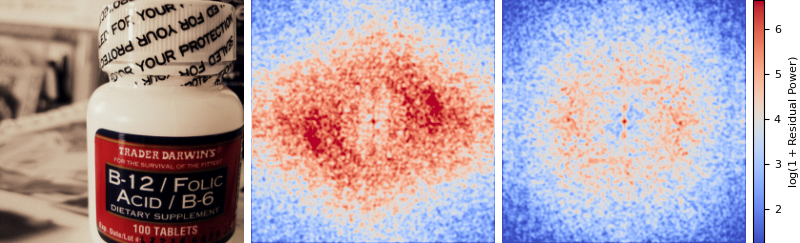}
  %  \caption*{\small Ours Spectrum (Example 1)}
\end{subfigure}
\hfill
\begin{subfigure}[t]{0.48\textwidth}
    \centering
    \includegraphics[width=\linewidth]{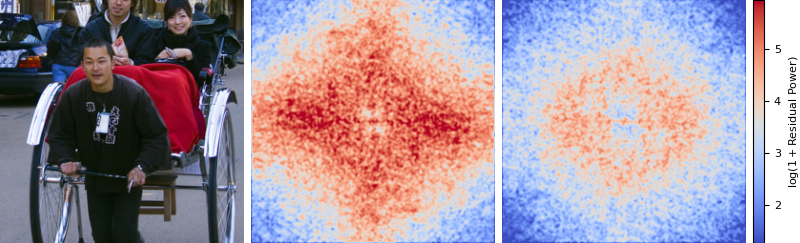}
    %\caption*{\small Ours Spectrum (Example 2)}
\end{subfigure}

\caption{Log residual power spectra of ImageNet validation images. Each set shows the input image (left), the residual power spectrum of VA-VAE (middle), and that of DeBaT (right). The residual power spectrum represents the Fourier-domain energy of the reconstruction error, where lower values indicate better reconstruction. Compared with VA-VAE, DeBaT consistently exhibits lower residual energy across both low- and high-frequency bands, demonstrating improved preservation of fine details, textures, and overall frequency-aware reconstruction.
}
\label{fig:two_examples_spectrum}
\end{figure*}

\section{Qualitative Comparisons} \label{sec:qualitative}

\subsection{Qualitative Comparison with Wavelet-based Tokenizers}

We qualitatively compare DeBaT reconstruction with wavelet-based latent VAE tokenizer, namely WF-VAE~\cite{li2025wfvae} and LiteVAE~\cite{sadat2024litevae} (see Figure 2 in the main paper). The goal of this comparison is to highlight the benefit of the proposed frequency-decoupled tokenizer. By explicitly separating frequency band and learning dedicated latent code for both low-frequency and high-frequency component, DeBaT improves reconstruction quality.

\autoref{fig:wave_comparisons} present qualitative reconstruction comparison. DeBaT produce more faithful reconstruction than WF-VAE and LiteVAE, with improved preservation of structural detail and fine texture. This improvement is also reflected in the corresponding error map, where DeBaT exhibit lower reconstruction error magnitude compared with baseline method.

It is important to note that WF-VAE was originally designed as a tokenizer for video reconstruction. To enable fair comparison in our setting, we adapt WF-VAE to the image modality and train it with comparable training budget as DeBaT. Specifically, we adapt WF-VAE to the image modality by training the model with single-frame inputs (T=1) using the image dataset loader. In this setting, the original 3D convolutional architecture operates over a temporal dimension of length one. The model is trained using the distributed training pipeline with the same optimization hyperparameters as the original implementation with batch size of 16 for 200k iterations.

\begin{figure*}[t]
\centering

% ==========================================================
% LEFT BLOCK (High Frequency)
% ==========================================================
\begin{minipage}{0.48\textwidth}
\centering

% High-frequency input (3 examples)
\begin{subfigure}{0.32\linewidth}
    \centering
    \includegraphics[width=\linewidth]{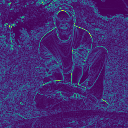}
    \caption*{\scriptsize HF Input}
\end{subfigure}
\begin{subfigure}{0.32\linewidth}
    \centering
    \includegraphics[width=\linewidth]{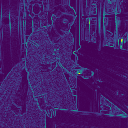}
    \caption*{\scriptsize  HF Input}
\end{subfigure}
\begin{subfigure}{0.32\linewidth}
    \centering
    \includegraphics[width=\linewidth]{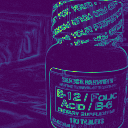}
    \caption*{\scriptsize  HF Input}
\end{subfigure}

\vspace{0.25cm}

% HF VA-VAE reconstructions
\begin{subfigure}{0.32\linewidth}
    \centering
    \includegraphics[width=\linewidth]{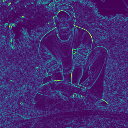}
    \caption*{\scriptsize VA-VAE  \hspace{1cm} RE: 0.0078}
\end{subfigure}
\begin{subfigure}{0.32\linewidth}
    \centering
    \includegraphics[width=\linewidth]{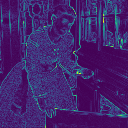}
    \caption*{\scriptsize VA-VAE \hspace{1cm} RE: 0.0054}
\end{subfigure}
\begin{subfigure}{0.32\linewidth}
    \centering
    \includegraphics[width=\linewidth]{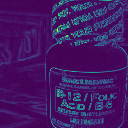}
    \caption*{\scriptsize VA-VAE \hspace{1cm} RE: 0.0061}
\end{subfigure}

\vspace{0.25cm}

% HF DeBaT reconstructions
\begin{subfigure}{0.32\linewidth}
    \centering
    \includegraphics[width=\linewidth]{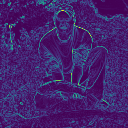}
    \caption*{\scriptsize DeBaT \hspace{1cm} RE: 0.0033}
\end{subfigure}
\begin{subfigure}{0.32\linewidth}
    \centering
    \includegraphics[width=\linewidth]{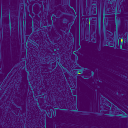}
    \caption*{\scriptsize DeBaT \hspace{1cm} RE: 0.0013}
\end{subfigure}
\begin{subfigure}{0.32\linewidth}
    \centering
    \includegraphics[width=\linewidth]{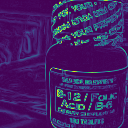}
    \caption*{\scriptsize DeBaT \hspace{1cm} RE: 0.0019}
\end{subfigure}

\end{minipage}
\hfill
% ==========================================================
% RIGHT BLOCK (Low Frequency)
% ==========================================================
\begin{minipage}{0.48\textwidth}
\centering

% Low-frequency input (3 examples)
\begin{subfigure}{0.32\linewidth}
    \centering
    \includegraphics[width=\linewidth]{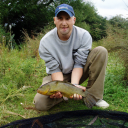}
    \caption*{\scriptsize LF Input}
\end{subfigure}
\begin{subfigure}{0.32\linewidth}
    \centering
    \includegraphics[width=\linewidth]{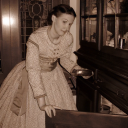}
    \caption*{\scriptsize LF Input}
\end{subfigure}
\begin{subfigure}{0.32\linewidth}
    \centering
    \includegraphics[width=\linewidth]{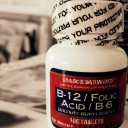}
    \caption*{\scriptsize LF Input}
\end{subfigure}

\vspace{0.25cm}

% LF VA-VAE reconstructions
\begin{subfigure}{0.32\linewidth}
    \centering
    \includegraphics[width=\linewidth]{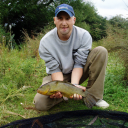}
    \caption*{\scriptsize VA-VAE \hspace{1cm} RE: 0.0016 }
\end{subfigure}
\begin{subfigure}{0.32\linewidth}
    \centering
    \includegraphics[width=\linewidth]{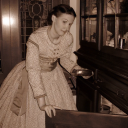}
    \caption*{\scriptsize VA-VAE \hspace{1cm} RE: 0.0006 }
\end{subfigure}
\begin{subfigure}{0.32\linewidth}
    \centering
    \includegraphics[width=\linewidth]{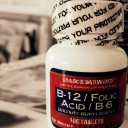}
    \caption*{\scriptsize VA-VAE \hspace{1cm} RE: 0.0017 }
\end{subfigure}

\vspace{0.25cm}

% LF DeBaT reconstructions
\begin{subfigure}{0.32\linewidth}
    \centering
    \includegraphics[width=\linewidth]{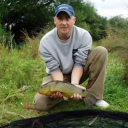}
    \caption*{\scriptsize DeBaT \hspace{1cm} RE: 0.0008 }
\end{subfigure}
\begin{subfigure}{0.32\linewidth}
    \centering
    \includegraphics[width=\linewidth]{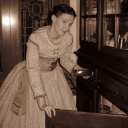}
    \caption*{\scriptsize DeBaT \hspace{1cm} RE: 0.0002 }
\end{subfigure}
\begin{subfigure}{0.32\linewidth}
    \centering
    \includegraphics[width=\linewidth]{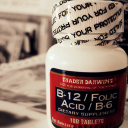}
    \caption*{\scriptsize DeBaT \hspace{1cm} RE: 0.0007 }
\end{subfigure}

\end{minipage}
\caption{ Visualization of high- and low-frequency inputs (HF \& LF Inputs) and their corresponding reconstructions. The left side shows the high-frequency inputs and their VA-VAE and DeBaT reconstructions, while the right side presents the analogous low-frequency components. DeBaT consistently achieves lower reconstruction error (RE) across both frequency bands, reflecting improved fidelity.}
\label{fig:low_high}
\end{figure*}

\begin{figure*}[t]
\centering

%%%%%%%%%%%%%%%%%%%%%%%%%%%%%%%%%%%%%%%%%%%%%%%%%%%%
%%%%%%%%%%%%%%%% Example 1 %%%%%%%%%%%%%%%%%%%%%%%%%
%%%%%%%%%%%%%%%%%%%%%%%%%%%%%%%%%%%%%%%%%%%%%%%%%%%%

\begin{subfigure}[b]{0.16\textwidth}
\centering
\textbf{\scriptsize Generation}\\[2pt]
\includegraphics[width=\linewidth]{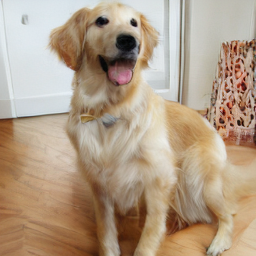}
\caption*{\scriptsize L2 Error:}
\end{subfigure}
\hfill
\begin{subfigure}[b]{0.16\textwidth}
\centering
\textbf{\scriptsize \scriptsize Train Image}\\[2pt]
\includegraphics[width=\linewidth]{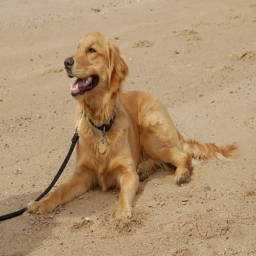}
\caption*{\scriptsize 0.2117}
\end{subfigure}
\hfill
\begin{subfigure}[b]{0.16\textwidth}
\centering
\textbf{\scriptsize Generation}\\[2pt]
\includegraphics[width=\linewidth]{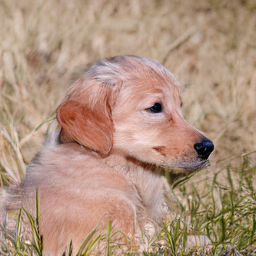}
\caption*{\scriptsize L2 Error:}
\end{subfigure}
\hfill
\begin{subfigure}[b]{0.16\textwidth}
\centering
\textbf{\scriptsize Train Image}\\[2pt]
\includegraphics[width=\linewidth]{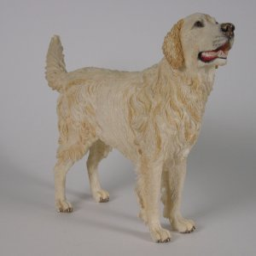}
\caption*{\scriptsize 0.1497}
\end{subfigure}
\hfill
\begin{subfigure}[b]{0.16\textwidth}
\centering
\textbf{\scriptsize Generation}\\[2pt]
\includegraphics[width=\linewidth]{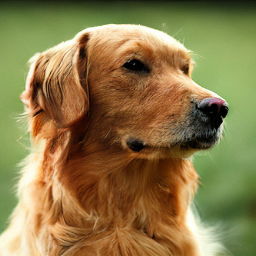}
\caption*{\scriptsize L2 Error:}
\end{subfigure}
\hfill
\begin{subfigure}[b]{0.16\textwidth}
\centering
\textbf{\scriptsize Train Image}\\[2pt]
\includegraphics[width=\linewidth]{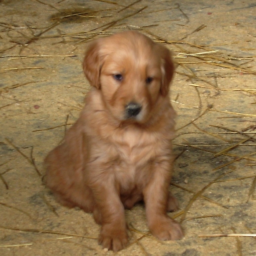}
\caption*{\scriptsize 0.2415}
\end{subfigure}

%%%%%%%%%%%%%%%%%%%%%%%%%%%%%%%%%%%%%%%%%%%%%%%%%%%%
%%%%%%%%%%%%%%%% Example 2 %%%%%%%%%%%%%%%%%%%%%%%%%
%%%%%%%%%%%%%%%%%%%%%%%%%%%%%%%%%%%%%%%%%%%%%%%%%%%%

\begin{subfigure}[b]{0.16\textwidth}
\centering
\includegraphics[width=\linewidth]{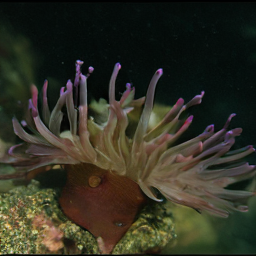}
\caption*{\scriptsize L2 Error:}
\end{subfigure}
\hfill
\begin{subfigure}[b]{0.16\textwidth}
\centering
\includegraphics[width=\linewidth]{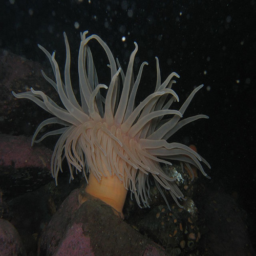}
\caption*{\scriptsize 0.1568}
\end{subfigure}
\hfill
\begin{subfigure}[b]{0.16\textwidth}
\centering
\includegraphics[width=\linewidth]{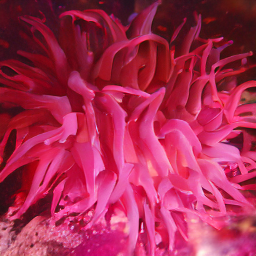}
\caption*{\scriptsize L2 Error:}
\end{subfigure}
\hfill
\begin{subfigure}[b]{0.16\textwidth}
\centering
\includegraphics[width=\linewidth]{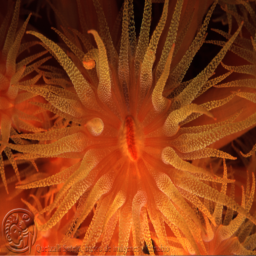}
\caption*{\scriptsize 0.2404}
\end{subfigure}
\hfill
\begin{subfigure}[b]{0.16\textwidth}
\centering
\includegraphics[width=\linewidth]{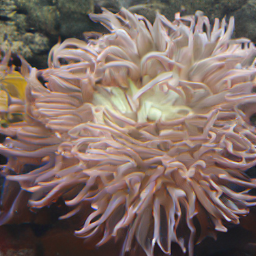}
\caption*{\scriptsize L2 Error:}
\end{subfigure}
\hfill
\begin{subfigure}[b]{0.16\textwidth}
\centering
\includegraphics[width=\linewidth]{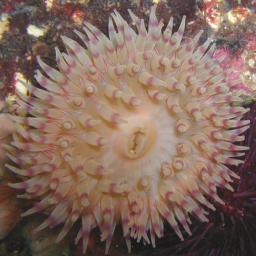}
\caption*{\scriptsize 0.1736}
\end{subfigure}

%%%%%%%%%%%%%%%%%%%%%%%%%%%%%%%%%%%%%%%%%%%%%%%%%%%%
%%%%%%%%%%%%%%%% Example 3 %%%%%%%%%%%%%%%%%%%%%%%%%
%%%%%%%%%%%%%%%%%%%%%%%%%%%%%%%%%%%%%%%%%%%%%%%%%%%%

\begin{subfigure}[b]{0.16\textwidth}
\centering
\includegraphics[width=\linewidth]{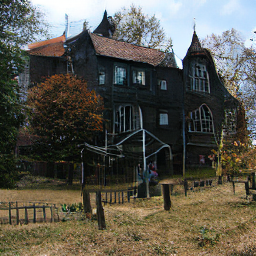}
\caption*{\scriptsize L2 Error:}
\end{subfigure}
\hfill
\begin{subfigure}[b]{0.16\textwidth}
\centering
\includegraphics[width=\linewidth]{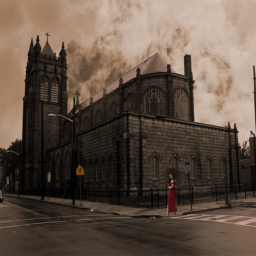}
\caption*{\scriptsize 0.2526}
\end{subfigure}
\hfill
\begin{subfigure}[b]{0.16\textwidth}
\centering
\includegraphics[width=\linewidth]{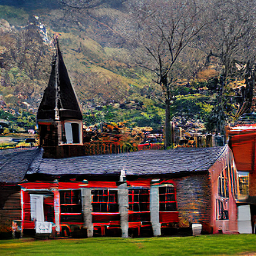}
\caption*{\scriptsize L2 Error:}
\end{subfigure}
\hfill
\begin{subfigure}[b]{0.16\textwidth}
\centering
\includegraphics[width=\linewidth]{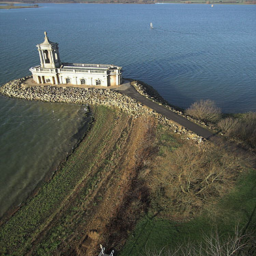}
\caption*{\scriptsize 0.2325}
\end{subfigure}
\hfill
\begin{subfigure}[b]{0.16\textwidth}
\centering
\includegraphics[width=\linewidth]{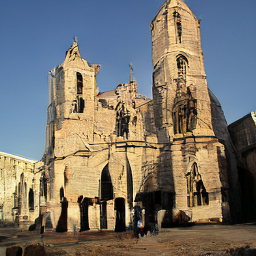}
\caption*{\scriptsize L2 Error:}
\end{subfigure}
\hfill
\begin{subfigure}[b]{0.16\textwidth}
\centering
\includegraphics[width=\linewidth]{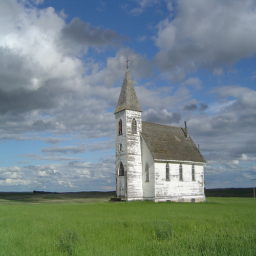}
\caption*{\scriptsize 0.2558}
\end{subfigure}

\caption{Class-conditional generation results (classes :golden retriever, sea anemone, castle)
of DeBaT. The generation column shows generated samples, while the corresponding training image column shows the nearest training image from training set for generation on left using normalized $\ell_2$ distance. The generated images are diverse.}
\label{fig:gen}

\end{figure*}

\subsection{Frequency-Domain Error Analysis}

We further analyze reconstruction behavior using the log residual power spectrum. The residual spectrum measures the Fourier-domain energy of the reconstruction error, revealing how much information the model fails to reproduce at each spatial frequency. In the visualization, \textcolor{red}{red} indicates higher residual energy (larger reconstruction error), while \textcolor{blue}{blue} indicates lower residual energy. The center of the spectrum corresponds to low-frequency components that capture global structure, whereas regions farther from the center correspond to high-frequency components responsible for fine details and textures.

\autoref{fig:wave_comparisons} shows the input image (left), the residual spectrum of VA-VAE (middle), and that of DeBaT (right). Across examples, VA-VAE exhibits higher residual energy across both low- and high-frequency regions, indicating larger reconstruction error throughout the spectrum. In contrast, DeBaT consistently suppresses residual energy, particularly in high-frequency regions, demonstrating improved preservation of fine structures and textures.

We compare DeBaT with VA-VAE, a strong pixel-wise spatial tokenizer, to contrast spatial tokenization with frequency-aware tokenization. The results highlight the benefit of explicitly modeling frequency bands in DeBaT, leading to more accurate frequency-aware reconstruction. Additionally, \autoref{fig:low_high} shows reconstruction errors for low- and high-frequency inputs, further demonstrating that DeBaT reconstructs both frequency components more accurately than VA-VAE.

\subsection{Qualitative Generation Evaluation}

We analyze class-conditional generation behavior of DeBaT by comparing generated samples with the nearest training image from the training set using normalized $\ell_2$ distance. \autoref{fig:gen} shows examples from three classes: \textit{golden retriever}, \textit{sea anemone}, and \textit{castle}. For each generated image, we retrieve the closest training image based on normalized $\ell_2$ similarity to assess whether the model memorizes training samples. Across examples, the generated samples exhibit noticeable variation from the nearest training images despite belonging to the same semantic class. The normalized $\ell_2$ distances further indicate that the generated images are not near-duplicates of training examples. These observations suggest that DeBaT learns meaningful class-conditional representations and generates diverse samples rather than memorizing training data.

\end{appendix}

\bibliographystyle{splncs04}
\bibliography{main}

\end{document}